\documentclass[journal]{IEEEtran}

\usepackage{cite}

\ifCLASSINFOpdf
\else
\fi
\usepackage{url}

\usepackage{graphicx}
\usepackage{color}
\usepackage{bbm}
\usepackage{mathtools}
\usepackage{ifthen}
\usepackage{subcaption}
\usepackage{amsfonts}
\usepackage{amsmath}
\usepackage{amssymb}
\usepackage{amsthm}
\usepackage{nicefrac}
\usepackage{xcolor}
\usepackage[normalem]{ulem}
\usepackage{url}
\DeclarePairedDelimiter{\ceil}{\lceil}{\rceil}
\DeclarePairedDelimiter{\floor}{\lfloor}{\rfloor}

\usepackage{algorithm}
\usepackage{algpseudocode}

\newtheorem{theorem}{Theorem}
\newtheorem{corollary}{Corollary}[theorem]
\newtheorem{lemma}{Lemma}
\newtheorem{definition}{Definition}

\newcommand{\cvargen}{\hat{c}_{n,\alpha}^{\dagger}}
\newcommand{\cvarte}{\hat{c}_{n,\alpha}^{(b)}}
\newcommand{\valp}{v_{\alpha}}
\newcommand{\calp}{c_{\alpha}}

\newcommand{\calpn}{\hat{c}_{n,\alpha}}

\newcommand{\calpgen}[1]{\overline{c}_{#1, \alpha}}
\newcommand{\meangen}[1]{\overline{\mu}_{#1}}

\newcommand{\knb}{K_{n, \beta}}
\newcommand{\pbb}{\mathbb{P}}
\newcommand{\cnb}{\ceil{n\beta}}
\newcommand{\fnb}{\floor{n\beta}}
\newcommand{\ksg}{k^{*}_{\gamma}}
\newcommand{\tea}{\hat{\mu}^{\dagger}}

\newcommand{\klest}{\widehat{\text{KL}}}
\newcommand{\obj}{\texttt{obj}}

\newcommand{\prob}[1]{\mathbb{P}\left( #1 \right)}

\newcommand{\calpni}[1]{\hat{c}_{N,\alpha,#1}}
\newcommand{\remove}[1]{}

\newcommand{\ra}{\rightarrow}

\newcommand{\da}{\downarrow}
\newcommand{\Exp}[1]{\mathbb{E}\left[#1\right]}
\newcommand{\Ind}[1]{\mathbbm{1}\left\{#1\right\}}

\newcommand{\M}{\mathcal{M}} 
 
\newcommand{\ignore}[1]{}
\newcommand{\nn}{\nonumber}

\newboolean{showcomments}
\setboolean{showcomments}{true}
\newcommand{\ak}[1]{  \ifthenelse{\boolean{showcomments}}
{ \textcolor{red}{(AK says:  #1)}} {}  }
\newcommand{\jk}[1]{  \ifthenelse{\boolean{showcomments}}
{ \textcolor{red}{(JK says:  #1)}} {}  }
\newcommand{\kj}[1]{  \ifthenelse{\boolean{showcomments}}
{ \textcolor{red}{(KJ says:  #1)}} {}  }
\newcommand{\addcites}[0]{\ifthenelse{\boolean{showcomments}}
{ \textcolor{green}{(add citation(s))}}{}}
\newcommand{\addref}[0]{\ifthenelse{\boolean{showcomments}}
{ \textcolor{green}{(add ref)}}{}}

\newcommand{\revision}[1]{{#1}}
\newcommand{\revisiontwo}[1]{{#1}}

\allowdisplaybreaks[4]
\begin{document}
\title{Statistically Robust, Risk-Averse Best Arm Identification in Multi-Armed Bandits}
%
%
%

\author{Anmol~Kagrecha, Jayakrishnan~Nair,
  and~Krishna~Jagannathan
  \thanks{A.~Kagrecha was with the Department of Electrical Engineering,
  IIT Bombay, Mumbai 400076, India. He is now with the Department of 
  Electrical Engineering, Stanford University, Stanford 94305, CA, USA.   
  J.~Nair is with the Department of
    Electrical Engineering, IIT Bombay, Mumbai 400076,
    India. K.~Jagannathan is with the Department of Electrical
    Engineering, IIT Madras, Chennai, 600036,
    India. Email:akagrecha@gmail.com, jayakrishnan.nair@iitb.ac.in,
    krishnaj@iitm.ac.in.}
  \thanks{A preliminary version of this paper was presented at Neural
    Information Processing Systems (NeurIPS) 2019
    \cite{Kagrecha2019}.}
  \thanks{Copyright (c) 2017 IEEE. Personal use of this material is permitted.  
  However, permission to use this material for any other purposes must be 
  obtained from the IEEE by sending a request to pubs-permissions@ieee.org.}}

\maketitle

\begin{abstract}
Traditional multi-armed bandit (MAB) formulations usually make certain assumptions about the underlying arms' distributions, such as bounds on the support or their tail behaviour. Moreover, such parametric information is usually `baked' into the algorithms. In this paper, we show that specialized algorithms that exploit such parametric information are prone to inconsistent learning performance when the parameter is misspecified. Our key contributions are twofold: (i) We establish fundamental performance limits of \emph{statistically robust} MAB algorithms under the fixed-budget pure exploration setting, and (ii) We propose two classes of algorithms that are asymptotically near-optimal. Additionally, we consider a risk-aware criterion for best arm identification, where the objective associated with each arm is a linear combination of the mean and the conditional value at risk (CVaR). Throughout, we make a very mild `bounded moment' assumption, which lets us work with both light-tailed and heavy-tailed distributions within a unified framework.
\end{abstract}

\begin{IEEEkeywords}
Multi-armed bandits, best arm identification, conditional value-at-risk,
concentration inequalities, robust statistics
\end{IEEEkeywords}

%
\IEEEpeerreviewmaketitle

\section{Introduction}
\label{sec:intro}
\IEEEPARstart{T}{he} multi-armed bandit (MAB) problem is fundamental
in online learning, where an optimal option needs to be identified
among a pool of available options. Each option (or arm) generates a
random reward/cost when chosen (or pulled) from an underlying unknown
distribution, and the goal is to quickly identify the optimal arm by
exploring all possibilities.

Classically, MAB formulations consider reward distributions with
bounded support, typically $[0,1].$ Moreover, the support is assumed
to be known beforehand, and this knowledge is baked into the
algorithm. However, in many applications, it is more natural to not
assume bounded support for the reward distributions, either because
the distributions are themselves unbounded (even heavy-tailed), or
because a bound on the support is not known {\it a priori}. There is
some literature on MAB formulations with (potentially) unbounded
rewards; see, for example, \cite{bubeck2013,vakili2013}. Typically, in
these papers, the assumption of a known bound on the support of the
reward distributions is replaced with the assumption that certain
bounds on the moments/tails of the reward distributions are
known. Additionally, some algorithms even require knowledge of a
lower bound on the sub-optimality gap between arms; see, for example,
\cite{yu2018}. However, such
prior information may not always be available. Even if available, it
is likely to be unreliable, given that moment/tail bounds are
typically themselves estimates based on limited data. Unfortunately,
the effect of the unavailability/unreliability of such prior
information on the performance of MAB algorithms has remained largely
unexplored in the literature.

As we show in this paper, the performance of MAB algorithms is quite
sensitive to the reliability of moment/tail bounds on arm
distributions that have been incorporated into them. Specifically, we
prove that such specialized algorithms can be \emph{inconsistent} when
presented with an MAB instance that violates the assumed moment
bounds. This motivates the design of \emph{statistically robust} MAB
algorithms, i.e., algorithms that guarantee consistency on \emph{any}
MAB instance. This requirement ensures that algorithms are robust to
misspecification of distributional parameters, and are not
`over-specialized' for a narrow class of parametrized instances.



Furthermore, the typical metric used to quantify the goodness of an
arm in the MAB framework is its expected return, which is a
risk-neutral metric. In some applications, particularly in finance,
one is interested in balancing the expected return of an arm with the
risk associated with that arm. This is particularly relevant when the
underlying reward distributions are unbounded, even heavy-tailed, as
is found to be the case with portfolio returns in finance; see
\cite{bradley2003}. In these settings, there is a non-trivial
probability of a `catastrophic' outcome, which motivates a risk-aware
approach to optimal arm selection.

In this paper, we seek to address the two issues described
above. Specifically, we consider the problem of identifying the arm
that optimizes a linear combination of the mean and the Conditional
Value at Risk (CVaR) in a fixed budget (pure exploration) MAB
framework. 
\revisiontwo{
Considering a mean-CVaR framework provides some flexibility to 
trade-off between risk-aversion and reward-seeking behavior.
It is similar in spirit to considering the popular mean-variance  
formulation. However, CVaR is a better-behaved risk metric
(see \cite{artzner1999}) compared to variance, 
in addition to having the same dimensional units as the mean loss. 
Moreover, the existence of CVaR requires only
the first moment to be well defined, while existence of variance
also requires the second moment to be well defined. This is 
important because we make very
mild assumptions on the arm distributions (the existence of a
$(1+\epsilon)$th moment for some~$\epsilon > 0$), allowing for
unbounded support and even heavy tails.
}
\ignore{
The CVaR is a classical metric used to capture the risk
associated with an option/portfolio; see \cite{artzner1999}.}
In this setting, our goal is design statistically robust algorithms
with provable performance guarantees.

Our contributions can be summarized as follows.
\begin{enumerate}
 \item We establish fundamental bounds on the performance of
  statistically robust algorithms. In the classical setting, where
  tail/moment bounds on the arm distributions are assumed to be known,
  it is possible to design specialized algorithms such that the
  probability of error for any instance that satisfies these bounds
  decays as $O(\mathrm{exp}(-\gamma' T)),$ where $\gamma' >0 $ is a
  constant that depends on the instance and $T$ is the budget of arm
  pulls; see \cite{yu2018, prashanth2019}. In contrast, we prove that
  it is impossible for statistically robust algorithms to guarantee an
  exponentially decaying probability of error with respect to the
  horizon~$T.$ This result highlights, on one hand, the `price' one
  must pay for statistical robustness. On the other hand, it also
  demonstrates the fragility of classical specialized
  algorithms---parameter misspecification can render them
  inconsistent.
  \item Next, we design two classes of statistically robust algorithms
  that are asymptotically near-optimal. Specifically, we show that by
  suitably scaling a certain function that parameterizes these
  algorithms, the probability of error can be made arbitrarily close
  to exponentially decaying with respect to the horizon.  In
  particular, the probability of error under our algorithms is the
  form $O(\mathrm{exp}(-\gamma T^{1-q})),$ where $\gamma > 0$ is an
  instance-dependent constant and $q \in (0,1)$ is an algorithm
  parameter. Another feature of our algorithms is that they are
  \emph{distribution oblivious}, i.e., they require no prior knowledge
  about the arm distributions. Our algorithms use sophisticated
  estimators for the mean and the CVaR, that are designed to work well
  with (highly variable) heavy-tailed arm distributions. Indeed, we
  show that the use the simplistic estimators based on empirical
  averages would result in an inferior power-law decay of the
  probability of error.
  \item We propose two novel estimators for the CVaR of (potentially)
  heavy-tailed distributions for use in our algorithms, and prove
  exponential concentration inequalities for these estimators; these
  estimators and the associated concentration inequalities may be of
  independent interest.
  \item While our proposed algorithms are \emph{distribution
  oblivious} as stated, we demonstrate that it is possible to
  incorporate noisy prior information about arm moment bounds into the
  algorithms without affecting their statistical robustness. Doing so
  improves the short-horizon performance of our algorithms over those
  instances that satisfy the assumed bounds, leaving the asymptotic
  behavior of the probability of error unchanged for \emph{all}
  instances.
\end{enumerate}

The remainder of this paper is organized as follows. A brief survey of
the related literature is provided below. We formally define the
formulation and provide some preliminaries in
Section~\ref{sec:formulation_prelims}. Fundamental lower bounds for
statistically robust algorithms are established in
Section~\ref{sec:lower_bound}. The design and analysis of the proposed
robust algorithms are discussed in Section~\ref{sec:dis_ob_alg}.
Numerical experiments are presented in Section~\ref{sec:experiments},
and we conclude in Section~\ref{sec:discussion}.


\subsection*{Related Literature}

There is a considerable body of literature on the multi-armed bandit
problem. We refer the reader to the books \cite{bubeck2012,lattimore2018}
for a comprehensive review. Here, we restrict ourselves to papers that
consider (i) unbounded reward distributions, and (ii) risk-aware arm
selection.

The papers that consider MAB problems with (potentially) heavy-tailed
reward distributions include:
\cite{bubeck2013,vakili2013,carpentier2014}, in which regret
minimization framework is considered, and \cite{yu2018}, in which the pure
exploration framework is considered. All the above papers take the expected return
of an arm to be its goodness metric. The papers \cite{bubeck2013,vakili2013}
assume prior knowledge of moment bounds and/or the suboptimality
gaps. The work \cite{carpentier2014} assumes that the arms belong to
parameterized family of distributions satisfying a second order Pareto
condition. The paper \cite{yu2018} does contain analysis of one 
distribution oblivious algorithm (see Theorem 2 in their paper). 
The oblivious approach considered there is based on 
empirical estimator for the mean and therefore, the performance 
guarantee derived there is much weaker than the lower bound; we 
elaborate on this in the Subsection~\ref{sec:empirical}. 

There has been some recent interest in risk-aware multi-armed bandit
problems. The setting of optimizing a linear combination of mean and 
variance in the regret minimization framework has been considered in
\cite{sani2012, vakili2016}. Use of the logarithm of the moment 
generating function of a random variable as the risk metric in a
regret minimization framework is studied in \cite{maillard2013} and
the learnability of general functions of mean and variance is studied
in \cite{zimin2014}.  In the pure exploration setting,
VaR-optimization has been considered
in \cite{david2016,david2018}. However, the CVaR is a more preferable
metric because it is a coherent risk measure (unlike the VaR); see
\cite{artzner1999}. Strong concentration results for VaR are available
without any assumptions on the tail of the distribution; see
\cite{kolla2019a}, whereas concentration results for CVaR are more
difficult to obtain. Assuming bounded rewards, the problem of
CVaR-optimization has been studied in \cite{galichet2013, tamkin2019}.
The paper \cite{cassel2018} looks at path dependent regret and
provides a general approach to study many risk metrics. In a recent
paper
\cite{prashanth2019}, CVaR optimization with heavy 
tailed distributions is considered, but prior knowledge of moment
bounds is also assumed. \revision{More recently,~\cite{agrawal2021}
considers the mean-CVaR objective in the fixed-confidence setting;
they assume knowledge of moment bounds on the arms and adapt
the \emph{track and stop} methodology pioneered by \cite{Garivier2016}
to devise an asymptotically (as the confidence parameter $\delta \da
0$) optimal algorithm.  They also propose novel estimators for the
CVaR and develop strong concentration inequalities for these
estimators. None of the above papers consider the problem of
risk-aware arm selection in a statistically robust manner, as is done
here. The only other work we are aware of that addresses statistical
robustness in the MAB context is our own recent
work~\cite{ashutosh2021}; this paper focuses on the related regret
minimization framework (in contrast, the present paper considers the
pure exploration fixed budget framework).}
\revisiontwo{
Subsequent to our conference paper \cite{Kagrecha2019}, there has been further work related to VaR and CVaR in the bandit setting. 
Thompson sampling based algorithms for CVaR based bandits are
explored in \cite{zhu2020} and \cite{baudry2021}. Identification of 
top $m$-arms with optimal VaR is considered in \cite{zhang2021} 
and a differentially private algorithm to identify arms with optimal
VaR is considered in \cite{nikolakakis2021}.   
}

\section{Problem Formulation and Preliminaries}
\label{sec:formulation_prelims}
In this section, we introduce some preliminaries, 
state our modeling assumptions, and formulate the problem of
risk-aware best arm identification.

\subsection{Preliminaries}

For a random variable $X,$ given a prescribed confidence level $\alpha
\in (0,1)$, the \emph{Value at Risk} (VaR) is defined as $\valp(X) =
\text{inf}(\xi: \pbb(X \leq \xi) \geq \alpha).$ If $X$ denotes the
loss associated with a portfolio, $\valp(X)$ can be interpreted as the
worst case loss corresponding to the confidence level $\alpha.$ The
\emph{Conditional Value at Risk} (CVaR) of $X$ at confidence level
$\alpha \in (0,1)$ is defined as
\begin{equation*}
\calp(X) = \valp(X) + \frac{1}{1-\alpha}\mathbb{E}[X-\valp(X)]^{+},
\end{equation*}
where $[z]^{+} = $ max$(0,z)$. Both VaR and CVaR are used extensively
in the finance community as measures of risk, though the CVaR is
often preferred as mentioned above. Typically, the confidence level
$\alpha$ is chosen between $0.95$ and $0.99$. Throughout this paper,
we use the CVaR as a measure of the risk associated with an arm. Let
$\beta := 1-\alpha.$ For the special case where $X$ is continuous with
a cumulative distribution function (CDF) $F_X$ that is strictly
increasing over its support, $\valp(X) = F_X^{-1}(\alpha).$ In this
case, the CVaR can also be written as $\calp(X) = \Exp{X|X \geq
  \valp(X)}$. Going back to our portfolio loss analogy, $\calp(X)$
can, in this case, be interpreted as the expected loss conditioned on
the `bad event' that the loss exceeds the VaR.

Next, we recall that the KL divergence (or relative entropy) between
two distributions is defined as follows. For two distributions $\rho$
and $\rho',$ with $\rho$ being absolutely continuous with respect to
$\rho'$,
\begin{equation*}
\text{KL}(\rho, \rho') := \int \log\left(
\frac{\text{d}\rho(x)}{\text{d}\rho'(x)} \right) \text{d}\rho(x).
\end{equation*}
\ignore{Throughout, we assume that the arm distributions satisfy the following
condition: A random variable $X$ is said to satisfy condition
\textbf{C1} if there exists $p > 1$ such that $\Exp{|X|^p} < \infty.$
Note that \textbf{C1} is only mildly more restrictive than assuming
the well-posedness of the MAB problem, which requires
$\Exp{|X|} < \infty$.}
Throughout, we assume that the arm distributions satisfy the following
condition: 
\revisiontwo{
\begin{definition}
A random variable $X$ is said to satisfy condition
\textbf{C1} if there exists $p > 1$ such that $\Exp{|X|^p} < \infty.$
\end{definition}
}
Note that \textbf{C1} is only mildly more restrictive than assuming
the well-posedness of the MAB problem, which requires
$\Exp{|X|} < \infty$.
In particular, all light-tailed distributions and most heavy-tailed
distributions used and observed in practice satisfy \textbf{C1}. 

\revision{An important class of heavy-tailed distributions that
  satisfy \textbf{C1} is the class of regularly varying distributions
  with index greater than~1 (see Proposition 1.3.6,
  \cite{bingham1989}). Formally, the complementary cumulative
  distribution function (c.c.d.f.) of a regularly varying random
  variable~$X$ with index~$\alpha$ satisfies $\overline{F}_X(x) =
  x^{-a}L(x),$ where $L(\cdot)$ is a slowly varying
  function.\footnote{\revision{The c.c.d.f. of the random variable~$X$
    is defined as $\overline{F}_X(x) := 1-F_X(x) = \prob{X>x}.$} A
  function $L: \mathcal{R}_+ \to \mathcal{R}_+$ is said to be slowly
  varying if $\lim_{x\to \infty} \frac{L(xy)}{L(x)} = 1$ for all
  $y>0.$} The class of regularly varying distributions is a
  generalization of the class of Pareto distributions, and are
  characterized by an \emph{asymptotically power-law tail}; in
  contrast, recall that the Pareto distribution is a precise
  power-law. Some examples of regularly varying distributions (other
  than the Pareto): the Student's t, Cauchy, Burr, Fr{\'e}chet and
  L\'{e}vy distributions. See Chapter~2 in \cite{NWZ-HTbook} for an
  accessible treatment of regularly varying distributions.}

\revision{Finally, we note that heavy tails (more specifically, power
  law tails) have been observed empirically in a wide range of
  contexts, including the distribution of wealth (recall the classical
  \emph{80-20 rule}, a.k.a., the \emph{Pareto principle}), extremal
  events in insurance and finance, Internet file sizes and word
  frequencies in language (see \cite{NWZ-HTbook} for a comprehensive
  overview). In the context of MABs, heavy-tailed arms arise in
  applications related to finance, networks, and queueing systems. In
  particular, there has been an interest in applying MAB and related
  frameworks for portfolio optimization (see \cite{Huo2017,Shen2015});
  portfolio/stock returns tend to be heavy-tailed
  (see~\cite{Cont2001}). Similarly, MAB algorithms have been applied
  to the optimization of the routing/scheduling policy in networks and
  queueing systems (see, for example, \cite{Gagliolo2011,Song2020});
  delays and processing times in such systems are often heavy-tailed
  (see, for example, \cite{Gomes2000,Park2000}).}

\ignore{Regularly varying distributions are used to model distribution of
wealth (Pareto's law), hard disk error rates (see
\cite{schroeder2010}), word frequencies (Zipf's law), among
others. Regularly varying distributions are characterized using the
complementary cumulative distribution function (CCDF) which is defined
as $\overline{F}(x) := 1 - F(x),$ where $F(\cdot)$ is a
distribution. The CCDF of a regularly varying random variable $X$ with
index~$a$ is of the form $\overline{F}_X(x) = x^{-a}L(x),$ where
$L(\cdot)$ is a slowly varying function.\footnote{A function $L:
\mathcal{R}_+ \to \mathcal{R}_+$ is said to be slowly varying if
$\lim_{x\to \infty} \frac{L(xy)}{L(x)} = 1$ for all $y>0.$} Some
examples of regularly varying distributions are the Student's
t-distribution, Cauchy distribution, and Pareto distribution
$(\overline{F}_X(x) = \left(\nicefrac{x_m}{x}\right)^a \mathbbm{1}\{x
\geq x_m\}).$ Regular variation essentially means that the tail of the
distribution decays polynomially slowly after a large enough value of
$x.$
}


\subsection{Problem Formulation}
Consider a multi-armed bandit problem with
$K$ arms, labeled
$1,2,\cdots,K.$ The loss (or cost) associated with
arm~$i$ is distributed as
$X(i),$ where it is assumed that all the arms satisfy {\bf
  C1}. Therefore, it follows that there exists $p \in (1,2],$ $B <
\infty,$ and $V < \infty$ such that
\begin{equation*}
  \Exp{|X(i)|^p} < B \text{ and } \Exp{|X(i) - \Exp{X(i)}|^p} < V \text{ for all }i.
\end{equation*}
We pose the problem as (risk-aware) loss minimization, which is of
course equivalent to (risk-aware) reward maximization.  Each time an
arm~$i$ is pulled, an independent sample distributed as
$X(i)$ is observed. Given a fixed budget of
$T$ arm pulls in total, our goal is to identify the arm that minimizes
$\obj(i) = \xi_{1} \Exp{X(i)} + \xi_{2}\calp(X(i)),$ where
$\xi_1$ and
$\xi_2$ are non-negative (and given) weights. This places us in the
\emph{fixed budget, pure exploration} framework. The performance of an
algorithm (a.k.a., policy) is captured by its probability of error,
i.e., the probability that it fails to identify an optimal arm. Note
that $(\xi_1,\xi_2) =
(1,0)$ corresponds to the classical mean minimization problem (see
\cite{audibert2010,yu2018}), whereas $(\xi_1,\xi_2) =
(0,1)$ corresponds to a pure CVaR minimization problem (see
\cite{galichet2013,prashanth2019}). Optimization of a linear
combination of the mean and CVaR has been considered before in the
context of portfolio optimization in the finance community (see \cite{salahi2013}),
but not, to the best of our knowledge, in the MAB framework. The performance
metric we consider is the probability of incorrect arm identification
(a.k.a., the probability of error).

We denote a bandit instance by the tuple
$\nu = (\nu_1, \cdots, \nu_K),$ where $\nu_i$ is the distribution
corresponding to $X(i)$ (that satisfies {\bf C1}). Let the space of
such bandit instances be denoted by $\M.$ The ordered values of the
objective are denoted as $\{\obj[i]\}_{i=1}^{K}$ where
$\obj[1] \leq \obj[2] \leq \cdots \leq \obj[K].$
The suboptimality gap $\Delta[i]$ is defined as the difference between
$\obj[i]$ and $\obj[1],$ i.e., $\Delta[i] = \obj[i] - \obj[1].$ Note
that the suboptimality gaps $\{\Delta[i]\}_{i=2}^{K}$ are ordered as
follows: $0 \leq \Delta[2] \leq \cdots \leq \Delta[K].$ The
probability of error for an algorithm $\pi$ on the instance
$\nu \in \mathcal{M}$ with a budget $T$ is denoted by
$p_e(\nu,\pi, T).$

Our focus in this paper is on \emph{statistically robust} algorithms.
Formally, we say an algorithm is statistically robust if it guarantees
\emph{consistency} over the space $\mathcal{M}.$ An algorithm $\pi$ is
said to be \emph{consistent} over the set $\widetilde{\M}$ of MAB
instances if, for any instance $\nu \in \widetilde{\mathcal{M}}$,
$\lim_{T \to \infty} p_e(\nu, \pi, T) = 0$ (see \cite{kaufmann2016}).
\revisiontwo{
As we discuss in Section~\ref{sec:lower_bound}, the inclusion of heavy-tailed 
distributions makes statistical robustness more challenging and this 
is the main focus of our paper.
}





In the following section, we explore the fundamental limits on the
performance of statistically robust algorithms.

\section{Fundamental performance limits for robust algorithms}
\label{sec:lower_bound}

In this section, we prove a fundamental lower bound on the performance
of any statistically robust algorithm. Specifically, we show that
there exists a class of MAB instances in $\M,$ such that any
statistically robust algorithm would have a probability of error that
decays \emph{slower than exponentially} with respect to the horizon
$T$ over those instances. In other words, it is impossible to
guarantee exponential decay of the probability of error with respect
to $T$ for robust algorithms. This is in sharp contrast to classical
specialized algorithms, which can offer such a guarantee (over the
narrow class of instances they are designed for).

To highlight this contrast, we begin by considering the classical
setting, where the algorithm is specialized to a restricted subset of
$\M.$ We first show that it is possible to construct bandit instances
such that any algorithm, even one that knows the distributions of the
arms up to a permutation, would have at least an exponentially decaying
probability of error with respect to $T.$ In the special case of mean
minimization, this result was proved in~\cite{audibert2010}. Here, we
extend the analysis to the case when the objective is a linear
combination of mean and CVaR.
\ignore{
\begin{theorem}
\label{lower-bound-non-obl-theorem}
Let $\nu_a$ and $\nu_b$ be Gaussian distributions with the same mean
$\mu$ but different variances $\sigma_a^2$ and
$\sigma_b^2> \sigma_a^2$. For any algorithm, the probability of error
for at least one of the bandit instances $\nu = (\nu_a, \nu_b)$ or
$\Tilde{\nu} = (\nu_b, \nu_a)$ satisfies:
\begin{equation*}
p_e \geq \frac{1}{4} \exp(-T(\text{KL}(\nu_b, \nu_a) + o(1)))
\end{equation*} 
where $o(1)$ term depends on $\sigma_a,$ $\sigma_b,$ and $T$ and tends
to zero as $T \ra \infty$.
\end{theorem}
}
\revision{
\ignore{
\begin{theorem}
\label{lower-bound-non-obl-theorem}
Let $K = 2,$ and consider an algorithm $\pi$ that is consistent over
$\M.$ For any bandit instance $\nu = (\nu_1, \nu_2)$ satisfying
$\obj(1) < \obj(2),$ 
\begin{align}
  \limsup_{T \to \infty} -\frac{1}{T} \log p_e(\nu,\pi, T) \leq \max (\text{KL}(\nu_1, \nu_2), 
  \text{KL}(\nu_2, \nu_1)).
\end{align}
\end{theorem}
}
\begin{theorem}
\label{lower-bound-non-obl-theorem}
Let $K = 2.$ Consider a bandit instance $\nu = (\nu_1, \nu_2) \in \M$
satisfying $\obj(1) \neq \obj(2),$ such that $\nu_1$ and $\nu_2$ are
mutually absolutely continuous. Any algorithm~$\pi$ that is consistent
over $\{(\nu_1,\nu_2),(\nu_2,\nu_1)\}$ satisfies
\begin{align*}
  \limsup_{T \to \infty} -\frac{1}{T} \log p_e(\nu,\pi, T) \leq \max
  (\text{KL}(\nu_1, \nu_2), \text{KL}(\nu_2, \nu_1)).
\end{align*}
\end{theorem}
The proof of Theorem~\ref{lower-bound-non-obl-theorem} can be found in
Appendix~\ref{app:lower-bound-non-obl}.}

It is also possible to construct specialized algorithms, which `know'
bounds on $(p,B,V)$ and/or $\Delta[2],$ that achieve an exponential
decay of the probability of error with respect to~$T,$ over all those
instances that satisfy these bounds. In the special cases of mean
minimization and CVaR minimization, such algorithms are proposed in
\cite{yu2018} and \cite{prashanth2019}, respectively.  Analogous
constructions can also be performed for the more general objective we
consider here, as we show in Section~\ref{sec:dis_ob_alg}.

We now turn to setting of statistically robust algorithms, which is
the primary focus of the present paper. Our main result, stated below,
shows that the fundamental performance limit for robust algorithms
differs considerably from that for specialized algorithms---it is
impossible to guarantee an exponentially decaying probability of error
in the oblivious setting. For simplicity, this result is stated for
the special case $K=2.$

\begin{theorem}
\label{fund-lower-bound-theorem}
Let $K = 2,$ and consider an algorithm $\pi$ that is consistent over
$\M.$ For any bandit instance $\nu = (\nu_1, \nu_2)$ satisfying
$\obj(1) < \obj(2),$ such that $\nu_1$ is a regularly
varying distribution with index $a>1$,
\begin{align}
\label{eq:lower_bound_final}
  \lim_{T \to \infty} -\frac{1}{T} \log p_e(\nu,\pi, T) = 0.
\end{align}
\end{theorem}
Note that the limit in \eqref{eq:lower_bound_final} captures the
exponential decay rate of $p_e(\nu,\pi, T)$ as $T \ra \infty;$ a value
of zero implies that $p_e(\nu,\pi, T)$ asymptotically decays slower
than exponentially.
It is also instructive that the instances for which this
'subexponential' decay is established involve heavy-tailed
(specifically, regularly varying) cost distributions. Indeed, the
impossibility result in Theorem~\ref{fund-lower-bound-theorem} holds
\emph{because} the class $\M$ of MAB instances of interest includes
instances with heavy-tailed arm distributions. If $\M$ were to be
restricted to \emph{light-tailed} arm distributions, then it can be
shown that the same impossibility result \emph{does not}
hold.\footnote{\revision{In particular, it is possible to devise
algorithms for which the probability of error decays exponentially for
all light-tailed instances (see~\cite{prashanth2019}).}}

\begin{figure}[htp]
\centering
\includegraphics[width=0.3\linewidth]{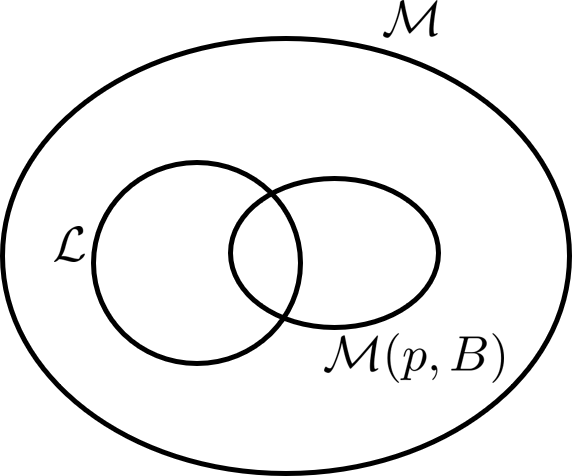}
\caption{Here, $\mathcal{L}$ refers to the class of MAB instances with
  light-tailed cost distributions. Since
  $\M(p,B) \setminus \mathcal{L}$ contains instances with regularly
  varying cost distributions, any algorithm that produces an
  exponentially decaying error probability over $\M(p,B)$ is
  necessarily not consistent over $\M.$}
\label{fig:distri_classes}
\end{figure}

Theorem~\ref{fund-lower-bound-theorem} also highlights the fragility
of classical specialized algorithms that have been proposed for
heavy-tailed instances. To see this, for $p > 1$ and $B > 0,$ let
$\M(p,B)$ denote the class of MAB instances where each arm
distribution lies in $\{\theta:\ \int |x|^{p} d\theta(x) \leq B\}.$
Note that $\M(p,B)$ contains both heavy-tailed as well as light-tailed
MAB instances; see Figure~\ref{fig:distri_classes}. As mentioned
before, it is possible to design algorithms that guarantee an
exponentially decaying probability of error over
$\M(p,B).$\footnote{This is done in \cite{yu2018} for the mean
minimization problem and in \cite{prashanth2019} for the CVaR
minimization problem.} Theorem~\ref{fund-lower-bound-theorem} implies
that such specialized algorithms are in fact \emph{not consistent}
over $\M.$ Indeed, if they were consistent, then their exponentially
decaying probability of error over the regularly varying instances in
$\M(p,B)$ would contradict Theorem~\ref{fund-lower-bound-theorem}. In
other words, while specialized algorithms perform very well over the
specific class of instances they are designed for, they necessarily
lose consistency over (certain) instances outside this class. In
practice, considering that moment bounds are themselves error prone
statistical estimates, Theorem~\ref{fund-lower-bound-theorem} shows
that specialized algorithms that exploit such bounds to provide strong
performance guarantees over the corresponding subset of bandit
instances are not robust to the inherent uncertainties in these
estimates.


The remainder of this section is devoted to the proof of
Theorem~\ref{fund-lower-bound-theorem}.
We note here that a similar impossibility result was proved by
\cite{glynn2015} for the pure exploration bandit problem in the fixed-confidence
setting. Our proof technique is inspired their methodology and also
relies crucially on the lower bounds
in~\cite{kaufmann2016}.\footnote{\revision{There are important
differences between the information theoretic lower bounds for the
fixed confidence setting and the fixed budget setting (compare, for
example, Theorems~4 and~12 in~\cite{kaufmann2016}. These differences
account for the contrasts between Theorem~1 in~\cite{glynn2015} and
Theorem~\ref{fund-lower-bound-theorem} in the present paper. In
particular, the former impossibility result is proved in the context
of light-tailed arm distributions having unbounded support, while the
latter is proved under instances containing heavy-tailed
(specifically, regularly varying) arms.}}
\ignore{
The key challenge in proving the impossibility result for the
fixed-budget setting is realizing that the complexity measures of the
fixed-confidence setting and the fixed-budget setting look similar at
the first glance but have some interesting differences . Considering
the class of all light-tailed distributions is sufficient to prove the
impossibility result in the fixed-confidence setting (compare, for
example, Theorems~4 and~12 in~\cite{glynn2015}).  However, for the
fixed-budget setting, a larger class of distributions has to be
considered; the class must contain heavy-tailed distributions.
Considering only unbounded distributions like in \cite{glynn2015} is
not sufficient for our case. The proof in \cite{glynn2015} has to be
carefully modified to get the impossibility result for our setting.}


We begin by stating a property of slowly varying functions $L(x)$ from
\cite[Proposition 1.3.6]{bingham1989}. 
\begin{lemma}
\label{slowly-varying-lemma}
If $L(\cdot)$ is a slowly varying function, then,
\begin{align*}
	\lim_{x\to\infty} x^\rho L(x) = \begin{cases}
		0 & \rho<0, \\
		\infty & \rho>0.
	\end{cases}
\end{align*}
\end{lemma}

The next lemma, which is a consequence of Theorem~12
in~\cite{kaufmann2016}, provides an information theoretic lower bound
on the rate of decay of the probability of error. While this result is
stated in \cite{kaufmann2016} for the classical mean optimization
problem, their arguments do not depend on the specific arm metric
used.
\begin{lemma}
\label{consistency-lemma}
Let $\nu = (\nu_1, \nu_2)$ be a two-armed bandit model such that
$\xi_1 \mu(1) + \xi_2 \calp(1) < \xi_1 \mu(2) + \xi_2 \calp(2)$
for given $\xi_1, \xi_2 \geq 0.$ Any consistent algorithm satisfies
\begin{equation}
\label{eq:lower_bound}
	\limsup_{t \to \infty} -\frac{1}{t} \log p_e(\nu, t) \leq c^*(\nu), 
\end{equation}
where, 
\begin{equation*}
	c^*(\nu) := \inf_{(\nu_1', \nu_2') \in \mathcal{M}: \obj'(1)>\obj'(2)}
	\max(\text{KL}(\nu_1', \nu_1), \text{KL}(\nu_2', \nu_2))
\end{equation*}	
\end{lemma} 

Next, we show that for any regularly varying distribution $F,$ one can
construct a perturbed distribution $G$ such that (i) $\text{KL}(G, F)$
is arbitrarily small, and (ii) the objective value $\obj(G)$ is
arbitrarily large. 
\ignore{
Lemma~\ref{slowly-varying-lemma} plays a key role in this
construction.  Intuitively, a $\rho<0$ in the distribution helps to
keep the KL divergence small while a $\rho>0$ in the mean and CVaR
helps to blow up the objective.}
\begin{lemma}
\label{lb-cond-lemma}
Consider a regularly varying distribution $F$ of index $p>1.$
Then given any $\delta \in (0,1)$ and $\gamma>\obj(F)$, there
exists a distribution $G,$ also regularly varying with
index~$p,$ such that
\begin{align*}
	&\text{KL}(G, F) \leq \delta, \\
	& \obj(G) \geq \gamma.
\end{align*}
\end{lemma} 

\begin{IEEEproof}
We have, using Lemma~\ref{slowly-varying-lemma},
\revision{ 
\begin{align*}
	\lim_{x\to\infty} x^\rho \overline{F}(x) = 
	\begin{cases}
	0 & \rho<p \\
	\infty & \rho>p
        \end{cases}.
\end{align*} \\
}
We construct the distribution $G$ as follows:
\begin{align*}
	G(x) &= \chi_{1} F(x) \text{ for } x<b \\
	\overline{G}(x) &= b^{p-0.5} \overline{F}(x) \text{ for } x\geq b
\end{align*}
where $b$ is a suitably large constant whose value we will set later.
We set \revision{$\chi_{1} = \frac{1-b^{p-0.5}\overline{F}(b)}{F(b)}$}
to ensure that $G(\cdot)$ is continuous at $b.$ As
$\lim_{b \to \infty} b^{p-0.5}\revision{\overline{F}(b)} = 0,$ we have
$0 < \chi_{1} < 1$ for large enough $b$, with
$\lim_{b \to \infty} \chi_{1} = 1.$ \revision{Note that under the
above construction, $G$ is also regularly varying with index~$p,$
though its tail (c.c.d.f.) is `heavier' than that of $F$ to the right
of~$b$ by a large multiplicative constant (also dependent on $b$). In
other words, the probability mass in~$G$ to the right of $b$ is
emphasized relative to~$F$, while the mass to the left of~$b$ is
correspondingly shrunk (by the factor~$\chi_{1}$). See
Figure~\ref{fig:lb_F_G} for a pictorial representation of the
probability density functions of $F$ and $G,$ assuming~$F$ has a
density.}
\begin{figure}[t]
\centering
\includegraphics[width=0.7\linewidth]{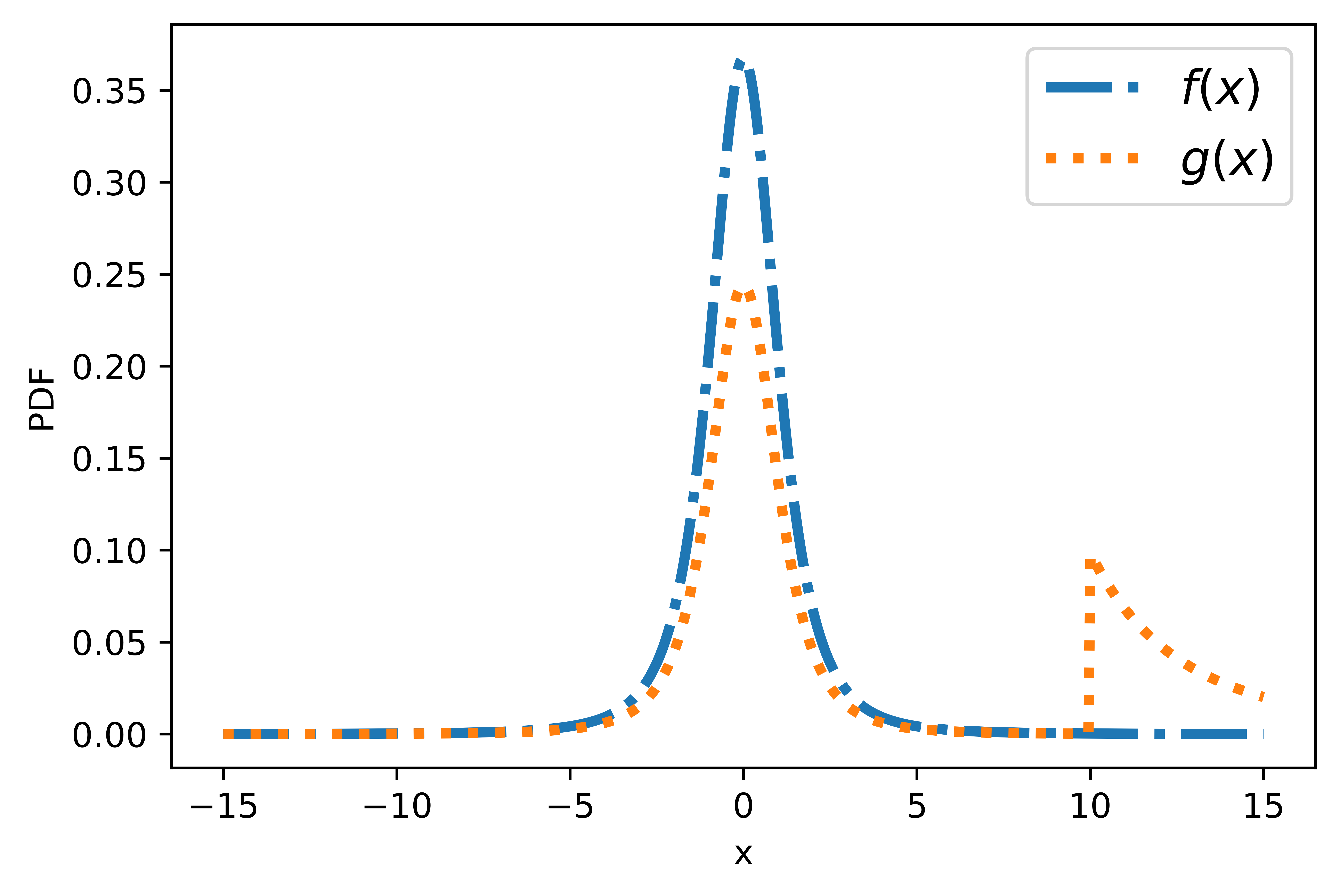}
\caption{Probability density functions~$f$ and $g$ corresponding
to the distributions~$F$ and $G,$ respectively, as per the
construction in the proof of Lemma~\ref{lb-cond-lemma}. Here, $F$ is a
standard Student's t-distribution with degrees of freedom parameter~3,
and $b = 10.$}
\label{fig:lb_F_G}
\end{figure}

The KL divergence between $G$ and $F$ is given by
\begin{align*}
	\text{KL}(G, F) &= \int_{-\infty}^{\infty} \log{\frac{dG(x)}{dF(x)}} dG(x) \\
	&= \int_{-\infty}^{b} \chi_{1} \log{\chi_{1}} dF(x) +  \int_{b}^{\infty} b^{p-0.5} \log{b^{p-0.5}} dF(x)\\
	&\leq b^{p-0.5}(p-0.5)\log(b) \revision{\overline{F}(b)} \quad (\because \chi_{1} < 1).
\end{align*}
As $\lim_{b \to \infty} b^{p-0.5} \log(b) \overline{F}(b) = 0$, we can
choose a large enough $b$ such that $\text{KL}(G, F) \leq \delta.$ We
further show that as $b$ tends to infinity, the mean and CVaR of $G$
also tend to infinity.  This ensures that for a suitably large $b,$
\obj(G) can be made greater than $\gamma.$

For $b$ such that $F(b^+) = F(b^-),$
\begin{align*}
\mu(G) &= \chi_{1} \int_{-\infty}^{b} x dF(x) + b^{p-0.5} \int_{b}^{\infty} x dF(x) \\
		&\geq \chi_{1} \int_{-\infty}^{b} x dF(x) + b^{p-0.5} \left(b\overline{F}(b)\right).
\end{align*}
As $\lim_{b \to \infty} b^{p+0.5} \overline{F}(b) = \infty$ and
$\lim_{b \to \infty} \chi_{1} \int_{-\infty}^{b} x dF(x) = \mu(F),$ we
have $\lim_{b \to \infty}\mu(G) = \infty.$

Similarly, for large enough $b$,
$\valp(G) = \inf(\xi: \chi_{1} F(\xi) \geq \alpha).$ Also,
$\lim_{b \to \infty} \valp(G) = \valp(F).$ For $b$ large enough such
that $F(b^+) = F(b^-),$
\begin{align*}
\calp(G) = &\frac{1}{1-\alpha}\int_{\valp(G)}^{b} \chi_{1} x dF(x) 
+ \frac{1}{1-\alpha} \int_{b}^{\infty} b^{p-0.5} x dF(x) \\
\geq &\underbrace{\frac{1}{1-\alpha}\int_{\valp(G)}^{b} \chi_{1} x dF(x)}_{T_1}
+ \underbrace{\frac{1}{1-\alpha} b^{p+0.5} \overline{F}(b)}_{T_2}.
\end{align*}
Note that $\lim_{b \to \infty}T_1 = \calp(F)$ and $\lim_{b \to \infty}T_2 = \infty.$
Hence, $\lim_{b \to \infty}\calp(G) = \infty.$ 
\end{IEEEproof}

Finally, Theorem~\ref{fund-lower-bound-theorem} is an immediate
consequence of Lemmas~\ref{consistency-lemma} and~\ref{lb-cond-lemma}.
\begin{IEEEproof}[Proof of Theorem~\ref{fund-lower-bound-theorem}]
  As $\nu_1$ is regularly varying, $\nu'_1$ can be chosen so that
  $KL(\nu'_1, \nu_1) \leq \delta$ for any small $\delta>0,$ and
  $\obj(\nu_1') > \obj(\nu_2) > \obj(\nu_1).$
  Considering the alternative instance $\nu' = (\nu'_1, \nu_2),$ an
  application of Lemma~\ref{consistency-lemma} implies that
  $c^*(\nu) \leq \delta.$ Since $\delta$ can be made arbitrarily
  small, it follows that $c^*(\nu) = 0.$ This proves
  Theorem~\ref{fund-lower-bound-theorem}.
\end{IEEEproof}


\section{Statistically robust algorithms}
\label{sec:dis_ob_alg}

In this section, we propose statistically robust, risk-aware
algorithms, and prove performance guarantees for these
algorithms. As enforced by the impossibility result proved in
Section~\ref{sec:lower_bound}, these algorithms produce an
(asymptotically) \emph{slower-than-exponential} decay in the
probability of error with respect to the budget $T.$ However, we show
that by tuning a certain function that parameterizes the estimators
used in these algorithms, the probability of error can be made
arbitrarily close to exponentially decaying. In this sense, the class
of algorithms proposed are \emph{asymptotically near-optimal}. This
is, however, not an entirely `free lunch'---tuning the algorithms to
be near-optimal asymptotically (as $T \ra \infty$) leads to a
potential degradation of performance for moderate values of~$T.$
Interestingly, if noisy prior information is available, say on moment
bounds satisfied by the arm distributions, this can be incorporated
into our algorithms to improve the short-horizon performance, without
affecting their statistical robustness.

This section is organised as follows. We begin by describing the basic
framework of the algorithms proposed here. In the following three
subsections, different algorithm classes are considered, along with
their corresponding performance guarantees. The different algorithm
classes differ only in the estimators used for the mean and CVaR of
each arm. Indeed, when dealing with heavy-tailed MAB instances, naive
estimators based on empirical averages perform poorly (as we
demonstrate in Section~\ref{sec:empirical}), necessitating the use of
more sophisticated estimators that are less sensitive to the
(relatively frequent) outliers that arise in heavy-tailed data (see
Sections~\ref{sec:truncated} and \ref{sec:median}).


Our algorithms are of \emph{successive rejects} (SR) type
\cite{audibert2010}. They are parameterized by positive integers
$n_1 \leq n_2 \leq \cdots \leq n_{K-1}$ satisfying $n_1 = \Omega(T)$
and $\sum_{i=1}^{K-2} n_i + 2 n_{K-1} \leq T.$ The algorithm proceeds
in $K-1$ phases, with one arm being rejected from further
consideration at the end of each phase. In phase~$i,$ the $K+1-i$ arms
under consideration are pulled $n_i-n_{i-1}$ times, after which the
arm with the worst (estimated) performance is rejected. This is
formally expressed in Algorithm~\ref{alg:gsr}. Here,
$\meangen{n_k}(i)$ and $\calpgen{n_k}(i)$ denote generic estimators of
the mean and CVaR of arm~$i$, respectively, using $n_k$ samples from
the corresponding distribution. The specific estimators used will
differ across the three classes of algorithms we describe later.

\revisiontwo{Note that the above framework, which we refer to as 
\emph{risk-aware generalized successive rejects}, allows for more 
than one arm elimination at once;
for example, if $n_i = n_{i+1},$ two arms (the worst performing, among
the surviving arms) would in effect be rejected after phase~$i.$ Most
well known algorithms for fixed budget MABs can be viewed as specific
instances of this risk-aware generalized successive rejects framework. For
example, the classical SR algorithm in \cite{audibert2010} used
$n_k \propto \frac{T-K}{K+1-k}.$ A related algorithm, called
\emph{sequential halving} (see~\cite{Karnin2013}),
eliminates half the surviving arms after each round of pulls (this
corresponds to $n_{1} = n_{2}=\cdots= n_{\frac{K}{2}},$
$n_{\frac{K}{2}+1} = n_{\frac{K}{2}+2} = \cdots = n_{\frac{3K}{4}},$
and so on). Another special case is \emph{uniform exploration} (UE),
where $n_1 = n_2 = \cdots n_{K-1} = \floor{T/K}.$ As the name
suggests, under uniform exploration, all arms are pulled an equal
number of times, after which the arm with the best estimate is
selected.}


\revisiontwo{
\begin{algorithm}
  \caption{Risk-aware generalized successive rejects algorithm}
  \label{alg:gsr}
\begin{algorithmic}
  \Procedure{RA-GSR}{$T, K, \{n_{1},\cdots,n_{K-1}\}$}
  \State $A_{1} \gets \{1,\cdots,K\}$
  \State $n_{0} \gets 0$
  \For{$k = 1 \text{ to }K-1$}
  \State $\text{For each } i \in A_{k}\text{, pull arm } i \text{ for } n_{k}-n_{k-1} \text{ rounds}$
  \State $\text{Let } A_{k+1} = A_{k} \setminus \text{arg max}_{i \in A_{k}} \xi_{1} \meangen{n_k}(i) 
  + \xi_{2} \calpgen{n_k}(i)$
  \EndFor
  \State $\text{Output unique element of } A_{K}$
  \EndProcedure
\end{algorithmic}
\end{algorithm}  
}
We note here that SR type algorithms require that the budget/horizon
$T$ be known a priori. However, if $T$ is not known a priori, any-time
variants can be constructed as follows: UE, implemented in a round
robin fashion is of course inherently any-time. 
\revisiontwo{Risk-aware generalized successive reject}
algorithms can also be made any-time using the well-known doubling
trick (see \cite{besson2018}). 


The probability of error of the risk-aware generalized successive
rejects algorithm can be upper bounded in the following manner. During phase
$k$, at least one of the
$k$ worst arms is surviving. Thus, if the optimal arm
$i^{*}$ is dismissed at the end of phase $k$, that means:
\begin{multline*}
  \xi_{1} \meangen{n_k}(i^{*}) + \xi_{2} \calpgen{n_k}(i^{*})  \geq \\
   \min_{i \in \{(K),(K-1),\cdots,(K+1-k)\}} \xi_{1} \meangen{n_k}[i] + \xi_{2} \calpgen{n_k}[i]
\end{multline*}
Using the union bound, we get:
\begin{align}
\begin{split}
p_{e} \leq &\sum_{k=1}^{K-1} \sum_{i=K+1-k}^{K} \mathbb{P} \Big(\xi_{1} \meangen{n_k}(i^{*}) + \xi_{2} \calpgen{n_k}(i^{*}) \\
&\qquad\qquad\qquad\qquad\qquad\geq \xi_{1} \meangen{n_k}[i] + \xi_{2} \calpgen{n_k}[i]\Big)    
\end{split} \nonumber\\
\begin{split}
=  &\sum_{k=1}^{K-1} \sum_{i=K+1-k}^{K} \mathbb{P} \Big( \xi_{1}(\meangen{n_k}(i^{*}) - \mu(i^{*}) - (\meangen{n_k}[i] - \mu[i] ) ) \\
&+ \xi_{2}(\calpgen{n_k}(i^{*}) - \calp(i^{*}) - (\calpgen{n_k}[i] - \calp[i]) ) \geq \Delta[i] \Big)  
\end{split} \nonumber \\
\begin{split}
\leq &\sum_{k=1}^{K-1} \sum_{i=K+1-k}^{K} \biggl(\prob{\xi_{1}(\meangen{n_k}(i^{*}) - \mu(i^{*})) \geq \Delta[i]/4} \\
&\qquad\qquad\quad+ \prob{\xi_{1}(\mu[i] - \meangen{n_k}[i] ) \geq \Delta[i]/4} \\
&\qquad\qquad\quad+ \prob{\xi_{2}(\calpgen{n_k}(i^{*}) - \calp(i^{*})) \geq \Delta[i]/4 } \\
&\qquad\qquad\quad+ \prob{\xi_{2}(\calp[i] - \calpgen{n_k}[i]) \geq \Delta[i]/4 } \biggr)
\end{split} \label{eq:alg_upper_bound}
\end{align}

The terms in the summation above can be bounded using suitable
concentration inequalities on the estimators
$\meangen{n}(\cdot)$ and
$\calpgen{n}(\cdot)$---these will be derived for the specific
estimators we use in the following subsections.


\subsection{Algorithms utilizing empirical average estimators}
\label{sec:empirical}
In this section, we consider the simplest oblivious estimators---those
based on empirical averages.
Unfortunately, these simple techniques do not enjoy good guarantees;
the probability of error decays \emph{polynomially} (i.e., as a
\emph{power law}) in $T.$ The fundamental reason for this is the poor
concentration properties of these estimators when the underlying
distribution is heavy-tailed.

We begin by stating the empirical CVaR estimator. We then state our
concentration inequality for this CVaR estimator, establish its
tightness, and point to analogous existing results for the empirical
mean estimator. Finally, we use these inequalities to show that the
probability of error of SR-type algorithms using these estimators
decays polynomially. This motivates the use of more sophisticated
estimators that provide stronger performance guarantees; this is the
agenda for the following two subsections.



Suppose that $\{X_{i}\}_{i=1}^{n}$ are $n$ IID samples distributed
as the random variable $X.$ Let $\{X_{[i]}\}_{i=1}^{n}$ denote the
order statistics of $\{X_{i}\}_{i=1}^{n}$ i.e.,
$X_{[1]} \geq X_{[2]} \cdots \geq X_{[n]}$. Recall that the classical
estimator for $\calp(X)$ given the samples $\{X_{i}\}_{i=1}^{n}$ (see
\cite{serfling2009}) is:
\begin{align*}
\calpn(X) = X_{[\ceil{n\beta}]} + 
\frac{1}{n\beta} \sum_{i=1}^{\floor{n\beta}}(X_{[i]} - X_{[\ceil{n\beta}]}).
\end{align*}
Now, we state the concentration inequality for $\calpn(X)$ when~$X$
satisfies {\bf C1}.
\begin{theorem}
\label{conv-cvar-ht-theorem}
Suppose that $\{X_{i}\}_{i=1}^{n}$ are IID samples distributed as
$X,$ where $X$ satisfies condition {\bf C1}. Given
$\Delta > 0,$
\begin{equation*}
  \prob{|\calpn(X) -  \calp(X)| \geq \Delta} \leq \frac{C(p, \Delta, V)}{n^{p-1}} 
  + o(\frac{1}{n^{p-1}}),
\end{equation*}
where $C(p, \Delta, V)$ is a positive constant.
\end{theorem}
The precise statement of the result, with explicit expressions for
$C(p, \Delta, V)$ and the $o(\frac{1}{n^{p-1}})$ term above can be
found in Appendix~\ref{app:conv-cvar-ht-proof}. Note that the upper
bound decays polynomially in $n.$ Contrast this with exponentially
decaying concentration bounds proved in \cite{wang2010} for bounded
random variables (see Lemma~\ref{cvar-bounded-lemma} below). The bound
in Theorem~\ref{conv-cvar-ht-theorem} in nearly tight in an order
sense, as shown in the following theorem. Similar upper and lower
bounds for the concentration of the empirical mean estimator are
provided in~\cite{bubeck2013}.

\begin{theorem}
\label{empirical-lower-bound-theorem}
Suppose that $\{X_{i} \}_{i=1}^n$ be IID samples distributed as $X,$
where $X\sim\mathrm{Pareto}(x_m,a),$ where $x_m > 0$ and
$a > 1.$\footnote{$X\sim\mathrm{Pareto}(x_m,a)$ means
  $P(X>x) = \frac{x_m^a}{x^a}$ for $x>x_m$ and 1 otherwise.} Then
\begin{align*}
  \pbb(\calpn(X) > \calp(X) + \Delta) \geq \frac{\beta x_m^a}{n^{a-1} (\calp(X)+\Delta)^a} 
  + o\left(\frac{1}{n^{a-1}}\right).
\end{align*}
\end{theorem}

The proof of Theorem~\ref{empirical-lower-bound-theorem} can be found
in Appendix~\ref{app:conv-cvar-lower-bound-proof}. If
$X\sim\mathrm{Pareto}(x_m,a),$ then $\Exp{X^{\theta}} < \infty$ for
$\theta < a$ and $\Exp{X^{\theta}} = \infty$ for $\theta \geq a.$
Thus, if $a \in (1,2],$ comparing the upper bound in
Theorem~\ref{conv-cvar-ht-theorem} to the lower bound in
Theorem~\ref{empirical-lower-bound-theorem}, $p < a$ and $p$ can be
made arbitrarily close to $a,$ suggesting the near-tightness of the
upper bound of Theorem~\ref{conv-cvar-ht-theorem}.

Theorem~\ref{conv-cvar-ht-theorem} and the analogous result for the
empirical mean estimator (see Lemma~3 in \cite{bubeck2013}), when
applied to \eqref{eq:alg_upper_bound}, imply that generalized SR
algorithms using empirical average based estimators have a probability
of error that is $O(\frac{1}{n^{p-1}}).$ Moreover,
Theorem~\ref{empirical-lower-bound-theorem} shows that this bound is
nearly tight in the order sense. We demonstrate this via the following
example.
\begin{corollary}
  Consider a two-arm instance $\nu = (\nu_1, \nu_2).$
  %
  Arm~1 is optimal and has a $\mathrm{Pareto}(x_m,a)$ distribution
  with $a>1.$ Arm~2 is a constant having a value such that
  $\obj(2) = \obj(1) + \Delta.$ The probability of error, $p_e$, of
  any SR-type algorithm using empirical estimators is bounded below by
  $\frac{C}{T^{a-1}} + o\left(\frac{1}{T^{a-1}}\right),$ where $C > 0$
  is an instance (and algorithm) dependent constant.
\end{corollary}
The above corollary is a direct consequence of
Theorem~\ref{empirical-lower-bound-theorem} and the analogous lower
bound for the concentration of the empirical mean from
\cite{bubeck2013}; the proof is omitted.

To summarize, SR algorithms using empirical estimators are
statistically robust, but exhibit poor performance, with a probability
of error that decays (in the worst case) polynomially with respect to
the horizon.

\subsection{Algorithms utilizing truncation-based estimators}
\label{sec:truncated}
In this section, we show that SR-type algorithms using
truncation-based estimators for the mean and CVaR have considerably
stronger performance guarantees compared to the power law bounds seen
in Section~\ref{sec:empirical}. Specifically, we show that by scaling
a certain truncation parameter as a suitably slowly growing function
of the budget $T,$ the probability of error for these algorithms can
be arbitrarily close to exponentially decaying in $T.$

In the following, we first propose a truncation based estimator for
CVaR, prove a concentration inequality for same, and finally evaluate
the performance of the SR-type algorithms that use these
truncation-based estimators.

\medskip
{\bf \noindent CVaR Concentration} 
\medskip

We begin by stating a concentration inequality for CVaR of bounded
random variables from \cite{wang2010}.
\begin{lemma}[Theorem~3.1 in \cite{wang2010}]
\label{cvar-bounded-lemma}
Suppose that $\{X_{i}\}_{i=1}^{n}$ are IID samples distributed as $X,$
where the support of $X,$ $\text{supp}(X) \subseteq [a,b].$ Then, for
any $\Delta \geq 0,$
\begin{equation*}
\prob{|\calpn(X) -  \calp(X)| \geq \Delta} \leq 6 \exp \left(-\frac{1}{11} n\beta 
\left(\frac{\Delta}{b-a} \right)^{2} \right).   	
\end{equation*}   
\end{lemma}

We now use Lemma~\ref{cvar-bounded-lemma} to develop a CVaR
concentration inequality for unbounded (potentially heavy-tailed)
distributions. In particular, our concentration inequality applies to
the following truncation-based estimator. For $b > 0,$ define
$$X^{(b)}_{i} = \min(\max(-b,X_i),b).$$ Note that $X_i^{(b)}$ is simply
the projection of $X_i$ onto the interval $[-b,b].$ Let
$\{X^{(b)}_{[i]}\}_{i=1}^{n}$ denote the order statistics of truncated
samples $\{X^{(b)}_{i}\}_{i=1}^{n}.$ Our estimator $\cvarte(X)$ for
$\calp(X)$ is simply the empirical CVaR estimator for $X^{(b)} :=
\min(\max(-b,X),b),$ i.e.,
\begin{align}
  \label{eq:CVaR-estimator}
  \cvarte(X) = \calpn(X^{(b)}) = 
  X_{[\ceil{n\beta}]}^{(b)} + \frac{1}{n\beta} \sum_{i=1}^{\floor{n\beta}}(X_{[i]}^{(b)} - X_{[\ceil{n\beta}]}^{(b)}).
\end{align}
A truncation-based estimator for the mean is well-known (see
\cite{bickel1965,bubeck2013}); it is given by
\begin{equation}
  \label{eq:mean-truncation}
  \tea_{n}(X) := \frac{\sum_{i = 1}^{n} X_{j} \Ind{|X_{j}| \leq b} }{n}.
\end{equation} 
Note that the nature of truncation performed for our CVaR estimator is
different from that in the truncation-based mean estimator, where
samples with an absolute value greater than~$b$ are set to zero.  In
contrast, our estimator projects these samples to the interval
$[-b,b].$ This difference plays an important role in establishing the
concentration properties of the estimator.

We are now ready to state the concentration inequality for
$\cvarte(X),$ which shows that the estimator works well when the
truncation parameter $b$ is large enough.
\begin{theorem}
\label{cvar-ht-theorem}
Suppose that $\{X_{i}\}_{i=1}^{n}$ are IID samples distributed as
$X,$ where $X$ satisfies condition {\bf C1}. Given
$\Delta > 0,$ 
\begin{equation}
\label{eq:cvar-ht-bound}
\prob{|\calp(X) - \cvarte(X)| \geq \Delta} \leq 6 \text{exp} 
\bigg(-n\beta\frac{\Delta^2}{176 b^2}\bigg)
\end{equation} 
\begin{equation}
  \label{eq:cvar-ht-truncation_bound}
  \text{ for } b > \max\left( |\valp(X)|,
    \left[\frac{2B}{\Delta\beta}\right]^{\frac{1}{p-1}} \right).
\end{equation}
\end{theorem}
\begin{IEEEproof}
\revision{ 
We begin by bounding the bias in CVaR resulting from our
  truncation. It is important to note that so long as
  $b > |\valp(X)|,$ $\valp(X) = \valp(X^{(b)}).$ Thus, for
  $b > |\valp(X)|,$
  \begin{align} 
    &|\calp(X) - \calp(X^{(b)})| \nn\\
    =~ &\calp(X) - \calp(X^{(b)}) \nn \\
    =~&\frac{1}{\beta}\bigg(\mathbb{E}[X\mathbbm{1}\{X \geq
    \valp(X)\}] - \mathbb{E}[X^{(b)} \mathbbm{1}\{X \geq \valp(X)\}]\bigg) \nn \\
    \stackrel{(a)}=~&\frac{1}{\beta}\mathbb{E}[(X-b) \mathbbm{1}\{X > b\}] \nn \\
    \leq~& \frac{1}{\beta}\mathbb{E}[X \mathbbm{1}\{X > b\}] \nn\\
    \stackrel{(b)}\leq~&
    \frac{B}{\beta b^{p-1}}. \label{eq:cvar_bias_bound}
\end{align}
}
Here, ($a$) is a consequence of $b > |\valp(X)|.$ The bound ($b$)
follows from
\begin{multline*}
\mathbb{E}[X \mathbbm{1}\{X > b\}] \leq
\Exp{\frac{X^p}{X^{p-1}}\mathbbm{1}\{X > b\}} \\
 \leq \frac{1}{b^{p-1}}
\Exp{|X|^p} \leq \frac{B}{b^{p-1}}.
\end{multline*}
It follows from \eqref{eq:cvar_bias_bound} that for $b$
satisfying~\eqref{eq:cvar-ht-truncation_bound}, the bias of our CVaR
estimator is bounded as:
$|\calp(X) - \calp(X^{(b)})| \leq \frac{\Delta}{2}.$ Thus, for $b$
satisfying \eqref{eq:cvar-ht-truncation_bound}, we have
\begin{align*}
  &\prob{|\calp(X) - \cvarte(X)| \geq \Delta} \\
  \leq~&\prob{|\calp(X) -
    \calp(X^{(b)})| + |\calp(X^{(b)}) - \calpn(X^{(b)})| \geq \Delta} \\
  \stackrel{(a)}\leq~& \prob{|\calp(X^{(b)}) - \calpn(X^{(b)})| \geq \frac{\Delta}{2}} \\
  \stackrel{(b)}\leq~& 6 \text{exp} \bigg(-n\beta
  \frac{(\Delta/b)^{2}}{176} \bigg).
\end{align*}
Here, ($a$) follows from the bound on $|\calp(X) - \calp(X^{(b)})|$
obtained earlier. For ($b$), we invoke Lemma~\ref{cvar-bounded-lemma}.
\end{IEEEproof}

In contrast with the concentration inequality for the empirical CVaR
estimator (see Theorem~\ref{conv-cvar-ht-theorem}), the
truncation-based estimator admits an \emph{exponential concentration
  inequality}. In other words, the probability of a $\Delta$-deviation
between the estimator and the true CVaR decays exponentially in the
number of examples, so long as the truncation parameter is set to be
large enough.

The key feature of truncation-based estimators like the one proposed
here for the CVaR is that they enable a parameterized bias-variance
trade-off. While the truncation of the data itself adds a bias to the
estimator, the boundedness of the (truncated) data limits the
variability of the estimator. Indeed, the condition that
$b > \left[\frac{2B}{\Delta \beta}\right]^{\frac{1}{p-1}}$ in the
statement of Theorem~\ref{cvar-ht-theorem} ensures that the estimator
bias induced by the truncation is at most $\Delta/2.$



However, in order to apply the proposed truncation-based estimator in
MAB algorithms, one must ensure that for each arm, the truncation
parameter satisfies the lower bound
\eqref{eq:cvar-ht-truncation_bound}. This is particularly problematic
in the context of statistically robust algorithms, which cannot
customize the truncation parameter to work for a narrow class of MAB
instances. Our remedy is to set the truncation parameter as an
increasing function of the number of data samples~$n,$ which ensures
that~\eqref{eq:cvar-ht-truncation_bound} holds for large enough~$n.$
Moreover, it is clear from~\eqref{eq:cvar-ht-bound} that for the
estimation error to (be guaranteed to) decay with~$n,$~$b^2$ can grow
at most linearly in~$n.$ Indeed, for our bandit algorithms, we
set~$b = n^{q},$ where~$q \in (0,1/2).$

Finally, we note that it is tempting to set $b$ in a
\emph{data-driven} manner, i.e., to estimate the VaR, moment bounds
and so on from the data, and set $b$ large enough so that
\eqref{eq:cvar-ht-truncation_bound} holds with high probability. The
issue however is that $b$ then becomes a (data-dependent) random
variable, and proving concentration results with such data-dependent
truncation is challenging.

\medskip
{\bf \noindent Performance Evaluation}
\medskip

We now evaluate the performance of SR-type algorithms using
truncation-based estimators for mean and CVaR. To simplify the
presentation, we present the results corresponding to the classical SR
algorithm of \cite{audibert2010}; here, $n_k
= \frac{T-K}{(K+1-k)\overline{\log}(K)},$ where $\overline{\log}(K) :=
1/2 + \sum_{i=2}^{K} 1/i$. \revision{Our results can easily be
generalized to other members of the class of risk-aware generalized SR
algorithms.} We will denote the truncation parameter for the CVaR
estimator as $b_c$ and the truncation parameter for the mean estimator
as $b_m.$ Specifically, in phase~$k$ of the algorithm, the mean
estimator, given by~\eqref{eq:mean-truncation}, uses the truncation
parameter $b_m(n_k) = n_k^{q_m}, q_m \in (0,1),$ whereas the CVaR
estimator, given by~\eqref{eq:CVaR-estimator}, uses truncation
parameter $b_c(n_k) = n_k^{q_c}, q_c \in (0,0.5).$

\ignore{
We estimate the performance of each arm as follows. Suppose that
arm~$i$ has been pulled $n$ times, and we observe samples
$X_1^i,X_2^i,\cdots,X_n^i.$ We use the following truncated empirical
estimator (see \cite{bickel1965,bubeck2013}) for the mean value
associated with the arm:
\begin{equation*}
  \tea_{n}(i) := \frac{\sum_{j = 1}^{n} X_{j}^i \Ind{|X_{j}^i| \leq b_m(n)} }{n},
\end{equation*} 
where $b_m(n) = n^{q_m}$ for $q_m \in (0,1).$ Note that we are growing
the truncation parameter $b_m$ sub-linearly in $n.$
Our estimator for the CVaR associated with arm~$i$ is the one
developed before, i.e.,
\begin{equation*}
  \cvargen = \hat{c}_{n,\alpha}^{(b_c(n))},
\end{equation*}
where $b_c(n) = n^{q_c}$ for $q_c \in (0,1/2).$

We consider the successive rejects algorithm. Let
}

\begin{theorem}
\label{sr-prob-of-error-tea}
Let the arms satisfy the condition \textbf{C1}.
The probability of incorrect arm identification for the successive
rejects algorithm using truncation based estimators is bounded as follows.
\begin{align*}
  p_{e} \leq & \sum_{i=2}^{K} (K+1-i) 2\text{exp}\bigg(-\frac{1}{16\xi_{1}}\Big(\frac{T-K}{\overline{\log}(K)}\Big)^{1-q_m} \frac{\Delta[i]}{i^{1-q_m}}\bigg) \\
        +&\sum_{i=2}^{K} (K+1-i) 6\text{exp} \bigg(-\frac{\beta}{2464 \xi_{2}^{2}}\Big(\frac{T-K}{\overline{\log}(K)}\Big)^{1-2q_c} \frac{\Delta[i]^{2}}{i^{1-2q_c}} \bigg)
\end{align*}
for $T> K+K\overline{\log}(K)n^{*},$ where
\begin{align*}
  n^{*} &= 
          \max \bigg(\Big(\frac{12\xi_{1}B}{\Delta[2]}\Big)^{\frac{1}{q_m\min(p-1,1)}}, 
          \Big(\frac{8\xi_{2}B}{\beta\Delta[2]}\Big)^{\frac{1}{q_c(p-1)}},\\
        & \qquad \qquad \qquad \qquad \Big(\frac{B}{\text{min}(\alpha,\beta) } \Big)^{\frac{1}{q_c p}} \bigg).   
  \end{align*}
\end{theorem}

The proof of Theorem~\ref{sr-prob-of-error-tea} can be found in
Appendix~\ref{app:sr-truncation}. Here, we highlight the main takeaways 
from this result.

First, note that the probability of error (incorrect arm
identification) decays to zero as $T \ra \infty,$ for any instance
in~$\M,$ meaning the proposed algorithm is statistically
robust. Moreover, as expected, \emph{the decay is slower than
  exponential in~$T;$} taking $q_m = q,$ $q_c = q/2$ for
$q \in (0,1),$ the probability of error is
$O(\mathrm{exp}(-\gamma T^{1-q}))$ for an instance dependent positive
constant $\gamma.$
Note that this bound on the probability of error is considerably
stronger than the power law bounds corresponding to algorithms that
use empirical estimators.

Second, our upper bounds only hold when $T$ is larger than a certain
instance-dependent threshold. This is because the concentration
inequalities on our truncated estimators are only valid when the
truncation interval is wide enough (to sufficiently limit the
estimator bias). As a consequence, our performance guarantees only
kick in once the horizon length is large enough to ensure that this
condition is met.

Third, there is a natural tension between the asymptotic behavior of
the upper bound for the probability of error and the threshold on $T$
beyond which it is applicable, with respect to the choice of
truncation parameters $q_m$ and $q_c.$ In particular, the upper bound
on $p_e$ decays fastest with respect to $T$ when $q_m,q_c \approx 0.$
However, choosing $q_m,q_c$ to be small would make the threshold on
the horizon to be large, since the bias of our estimators would decay
slower with respect to $T.$ Intuitively, smaller values of $q_m,q_c$
limit the variance of our estimators (which is reflected in the bound
for $p_e$) at the expense of a greater bias (which is reflected in the
threshold on $T$), whereas larger values of $q_m,q_c$ limit the bias
at the expense of increased variance.

Finally, we note that while the truncation-based SR algorithm as
stated is \emph{distribution oblivious} (i.e., it assumes no prior
information about the arm distributions), noisy prior information
about the arm distributions can be used to tailor the scaling of the
truncation parameters. For example, suppose that it is believed that
the MAB instance belongs to $\M(p,B)$ and that the suboptimality gaps
are bounded below by $\Delta$ (i.e., $\Delta[2] \geq \Delta$). A
natural choice for the truncation parameters would then be
\begin{align*}
  b_m &= \left(\frac{12B\xi_1}{\Delta}\right)^{\frac{1}{p-1}} +
        T^{q}, \\
  b_c &= \max\left(
        \left(\frac{B}{\beta}\right)^{\frac{1}{p}},
        \left[\frac{8B\xi_2}{\Delta\beta}\right]^{\frac{1}{p-1}} \right)
        + T^{q/2}, 
\end{align*}
for small $q \in (0,1)$; this would make $n^*$ close to zero for
instances in $\M(p,B)$ having sub-optimality gaps exceeding $\Delta,$
while ensuring that the probability of error remains
$O(\mathrm{exp}(-\gamma T^{1-q}))$ for any instance in~$\M.$
Essentially, our prescription on the use of noisy prior information
about the arm distributions is to set the truncation parameters as the
`specialized' value suggested by the prior information, plus a slowly
growing function of the horizon to ensure robustness to the
unreliability to the prior information.



\subsection{Algorithms utilizing median-of-bins estimators}
\label{sec:median}
In this section, inspired by the median-of-means estimator (see
\cite{alon1999, bubeck2013}), we propose a similar estimator for CVaR
and we call it the median-of-cvars estimator.  The idea of this
estimator is to divide the samples into disjoint bins, compute the
empirical CVaR estimator for each bin, and to finally use the median
of these estimates.
In the following, we first derive a concentration inequality for the
median-of-cvars estimator. We then use this result, in conjunction
with known concentration properties of the median-of-means estimator,
to characterize the performance of SR algorithms that utilise such
\emph{median-of-bins} estimators.

\medskip
{\bf \noindent CVaR Concentration} 
\medskip 

We are now ready the state the first result of this section, 
which is a concentration inequality for the median-of-cvars estimator.
\begin{theorem}
\label{cvar-mob-theorem}
Suppose that $\{X_{i}\}_{i=1}^{n}$ are IID samples distributed as $X,$
where $X$ satisfies condition {\bf C1}. Divide the sampled into $k$
bins, each containing $N = \floor{n/k}$ samples, such that bin~$i$
contains the samples $\{X_{j}\}_{j=(i-1)N+1}^{i N}\}.$ Let
$\calpni{i}$ denote the empirical CVaR estimator for the samples in
bin~$i.$ Let $\hat{c}_M$ denote the median of empirical CVaR
estimators $\{\calpni{i}\}_{i=1}^{k}.$ Then given $\Delta > 0,$
\begin{equation}
 	\prob{|\hat{c}_M - \calp(X)| \geq  \Delta} \leq \text{exp}(-\frac{n}{8N})\\
 \end{equation} 
 if $N \geq N^{*},$ where $N^*$ is a constant that depends on the
 distribution of $X$ and $\Delta.$
\end{theorem}

A precise characterization of the constant $N^{*},$ along with the
proof of Theorem~\ref{cvar-mob-theorem}, are provided in
Appendix~\ref{app:cvar-mob-proof}. Note that like in the case of the
truncation-based estimator, the median-of-cvars estimator admits an
exponential concentration inequality (so long as the number of samples
per bin exceeds the threshold~$N^*$).

The (distribution dependent) lower bound~$N^*$ on the number of
samples per bin ensures a minimal degree of reliability of the
empirical CVaR estimator for each bin. Since we are interested in
applying the median-of-cvars estimator in statistically robust
algorithms, ensuring that this condition is satisfied for all arms is
problematic. As before, our remedy is to set the number of samples per
bin as a (slowly) growing function of the horizon~$T,$ which ensures
that the condition for the CVaR concentration to become meaningful
holds so long as the horizon $T$ exceeds an instance-specific
threshold.

The median-of-means estimator $\hat{\mu}_M$ is computed in a similar
fashion, i.e., by taking the median of empirical mean estimators for
bins $\{\{X_{j}\}_{j=(i-1)N+1}^{i N}\}_{i=1}^{k}.$ A concentration
inequality similar to Theorem~\ref{cvar-mob-theorem} can be proved for
$\hat{\mu}_M$ (see~\cite{bubeck2013} or Appendix~\ref{app:sr-mob}).



\medskip
{\bf \noindent Performance Evaluation}
\medskip

\revision{As before, for simplicity, we present our results assuming that phase
lengths are set as per the SR algorithm of~\cite{audibert2010}; the
generalization to other SR-type algorithms (as described in
Algorithm~\ref{alg:gsr}) is straightforward.} For our statistically
robust algorithms, we scale the number of samples per bin as
follows. In phase~$k$ of successive rejects, we set the number of
samples per bin for the CVaR estimator as $N_c = n_k^{q_c},$ where
$q_c \in (0,1),$ and the number of samples per bin for the mean
estimator as $N_m = n_k^{q_m},$ for $q_m \in (0,1).$ We now state the
upper bound on the probability of error for this algorithm.
\begin{theorem}
\label{sr-prob-of-error-mob}
Let the arms satisfy the condition \textbf{C1}.
The probability of incorrect arm identification for the successive
rejects algorithm using median-of-bins estimators is bounded as follows
\begin{multline*}
	p_e \leq \sum_{k=1}^{K-1} k \Bigg[ \exp\left(-\frac{1}{8}
	\left(\frac{T-K}{\overline{\log}(K)(K+1-k)}\right)^{1-q_m} \right) \\ 
	+ \exp\left(-\frac{1}{8}
	\left(\frac{T-K}{\overline{\log}(K)(K+1-k)}\right)^{1-q_c} \right)
	\Bigg]
\end{multline*}
for $T>T^{*},$ where $T^*$ is a instance-dependent threshold.
\end{theorem}

The explicit expression for $T^*$ and the proof of the theorem can be
found in Appendix~\ref{app:sr-mob}. In the following, we highlight the
key takeaways from Theorem~\ref{sr-prob-of-error-mob}.

First, SR algorithms based on median-of-bins estimators are
statistically robust, like their truncation-based
counterparts. Indeed, setting $q_c = q_m = q,$ the probability of
error is $O(\mathrm{exp}(-\gamma T^{1-q}))$ for any instance in $\M,$
where $\gamma$ is a positive instance-dependent constant. In other
words, the probablity of error decays sub-exponentially, but much
faster than the power law decay arising from the use of empirical
averages.

Second, our performance guarantee only hold when $T$ is large
enough. As with the truncation-based approach, this is because
favourable concentration properties of the median-of-bins estimators
only apply when the horizon is large enough.

Third, there is again a tension between the bound on the probability
of error and the threshold on $T$ beyond which the bounds are
applicable, with respect to the choice of $q_m$ and $q_c$. To get the
best asymptotic upper bound, $q_m$ and $q_c$ should be close to zero,
but this would make $T^*$ large, affecting the the short-horizon
performance.


Finally, we note that the SR algorithm using median-of-bins estimators
as stated is also \emph{distribution oblivious}. However, as with the
truncation-based approach, noisy prior information about the instance
can be used to tailor the scaling of the bin sizes for mean and CVaR
estimation to improve the short-horizon performance. For example, if
it is believed that the MAB instance belongs to $\mathcal{M}(p,B,V)$
and the suboptimality gaps are bounded below by $\Delta,$ the mean and
CVaR bin sizes may be chosen as follows:
\begin{align*}
  N_m &= \frac{576 \xi_1 V}{\Delta} + T^{q},\\
  N_c &= N^* + T^{q}, 
\end{align*}
where $q \in (0,1)$ is small and $N^*$ is a constant that depends on
$(p,B,V,\Delta,\xi_2)$ (see Appendix~D for details). This choice would
make $T^*$ close to zero for instances that lie in the sub-class under
consideration, without affecting the overall statistical robustness of
the algorithm. As before, this choice boils down to the `specialized'
choice of bin size dictated by the moment bounds, plus a slowly
growing function of the horizon for robustness.

\section{Numerical Experiments}
\label{sec:experiments}
\begin{figure}[t]
    \centering
    \begin{subfigure}[b]{0.9\linewidth}
        \includegraphics[width=\linewidth]{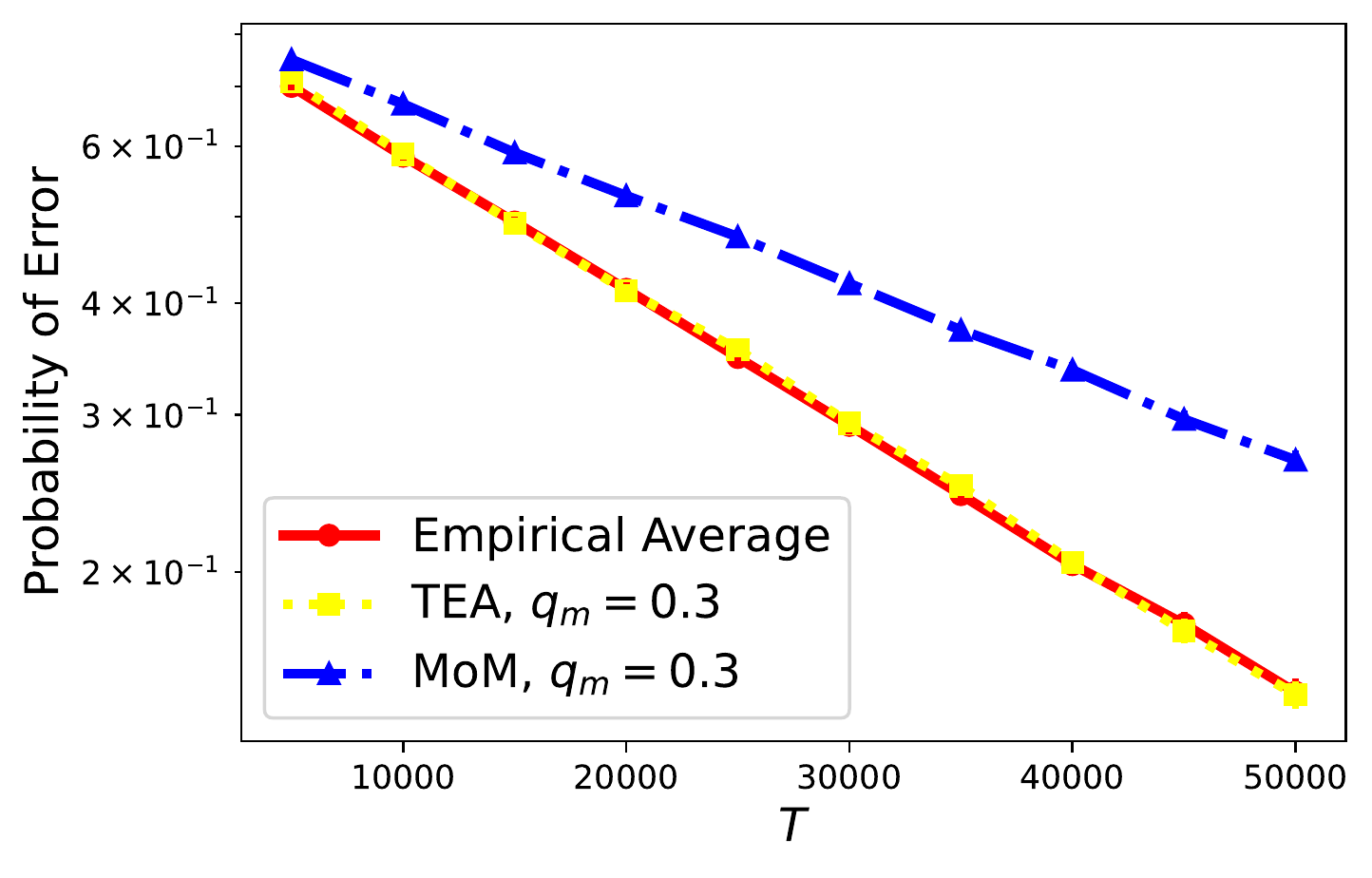}
        \caption{Mean Minimization}
        \label{fig:exp_mean}
    \end{subfigure} \\
    \begin{subfigure}[b]{0.9\linewidth}
        \includegraphics[width=\linewidth]{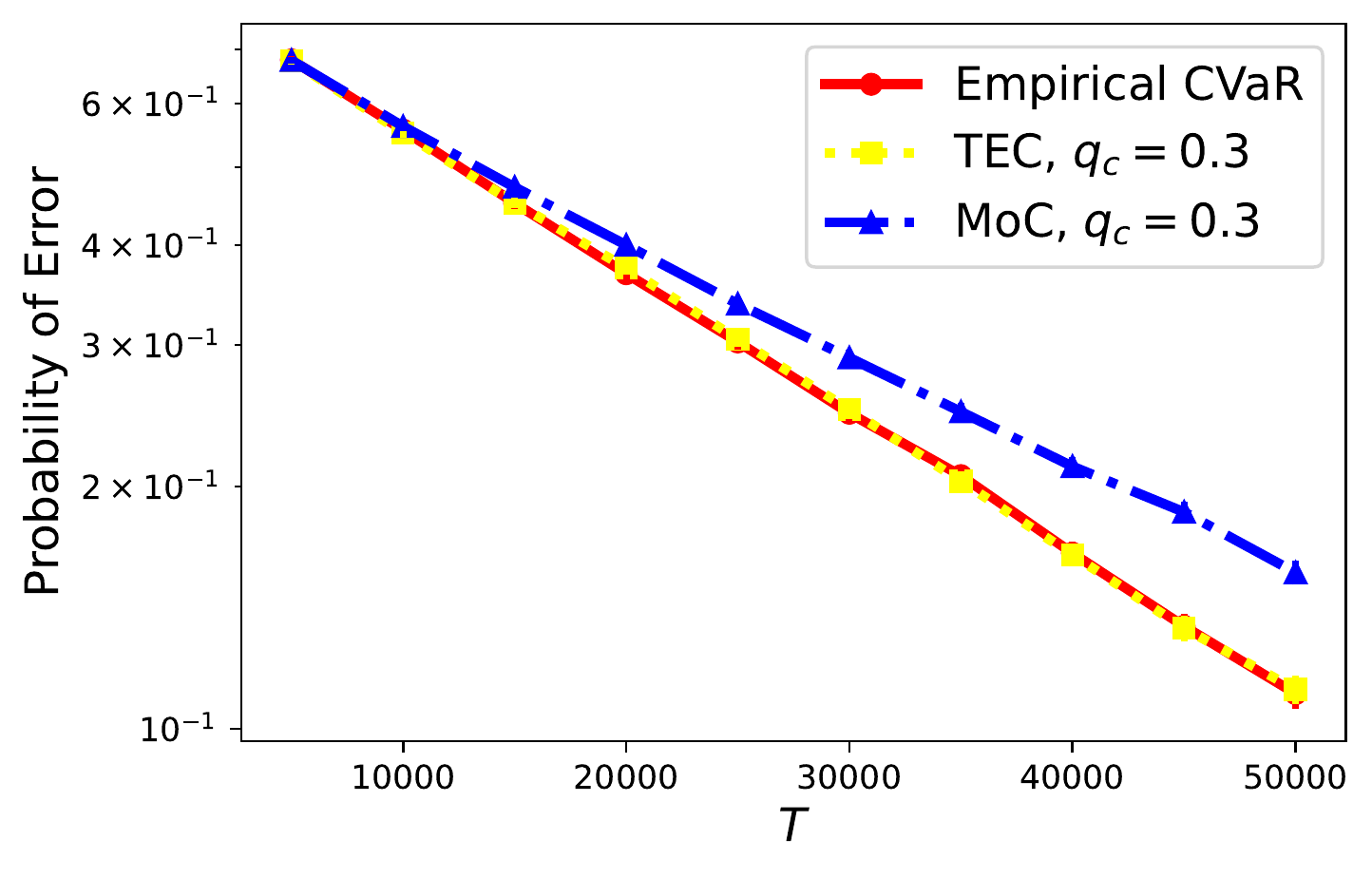}
        \caption{CVaR Minimization}
        \label{fig:exp_cvar}
    \end{subfigure}
    \label{fig:exp_case}
    \caption{Exponentially Distributed Arms}
\end{figure}

\ignore{
In this section, we evaluate the performance of the three
statistically robust algorithm classes presented in the previous
section. We also provide an example to demonstrate the fragility of
`specialized' based on noisy prior information.
We restrict ourselves to the classical successive rejects (SR) sizing
of phases, and two specific objectives: (i) mean minimization, i.e.,
$(\xi_1,\xi_2) = (1,0),$ and (ii) CVaR minimization, i.e.,
$(\xi_1,\xi_2) = (0,1).$ In each of the experiments below, the
probability of error is computed by averaging over 50000 runs at each
value of $T$. For CVaR minimization, the confidence level $\alpha$ is
set to 0.95.}
\revisiontwo{
In this section, we evaluate the performance of the three 
statistically robust algorithm classes presented in the previous
section. We also provide an example to demonstrate the fragility
of 'specialized' algorithms based on noisy prior information. 
We restrict ourselves to the classical successive rejects (SR) sizing
of phases, and primarily focus on two specific objectives: 
(i) mean minimization, i.e., $(\xi_1,\xi_2) = (1,0),$ and 
(ii) CVaR minimization, i.e., $(\xi_1,\xi_2) = (0,1).$ At the
end of this section, we also demonstrate the performance of 
our algorithms on two instances where the objective is a non-trivial
convex combination of mean and CVaR. In all the experiments below,
CVaR is calculated at a confidence level $\alpha$ of 0.95. 
Moreover, in each of the experiments below, the
probability of error is computed by averaging over 50000 runs at each
value of $T$. Confidence intervals for probability of error are 
calculated at a confidence of $99.9\%.$ As the number of runs is
quite large, the confidence intervals are not visible unless the
probabilities are very small.  
}

\noindent {\bf Light-tailed arms:} Consider the case when all the arms
are light-tailed. In particular, for mean minimization we consider the
following MAB problem instance: there are 10 arms, exponentially
distributed, the optimal having mean loss 0.97, and the remaining
having mean loss 1. For CVaR minimization, consider the following MAB
problem instance: there are 10 arms, exponentially distributed, the
optimal having a CVaR 2.85, and the remaining having a CVaR 3.00.
Parameters $q_m$ and $q_c$ for the truncation estimators (see
Section~\ref{sec:truncated}) are set to 0.3. Parameters $q_m$ and
$q_c$ for the median-of-bins estimators (see Section~\ref{sec:median})
are also set to 0.3.

As can be seen in Figure~\ref{fig:exp_mean} and
Figure~\ref{fig:exp_cvar}, the truncation-based algorithms and the
algorithms using empirical averages perform comparably well, whereas
median-of-bins algorithms produce an inferior performance. Because the
arm distributions have limited variability in this example, the
truncation-based estimators introduce very little bias, and are nearly
indistinguishable from the estimators based on empirical averages. On
the other hand, the median-of-bins estimators suffer from the poorer
concentration of the empirical averages per bin, which is not
sufficiently compensated by computing the median across
bins.\footnote{Indeed, this effect can be formalized in the special
  case of the exponential distribution.}


\ignore{
However, growing
the samples per bin as $n^{0.3}$ introduces too much bias and degrades
performance. Note that growing the number of samples per bin as $n^1$ 
or in other words, having just one bin reduces the median-of-bins estimators
to empirical estimators. Hence, in this case, having a few bins with very strong 
concentration gives better performance. } 




\begin{figure}[t]
    \centering
    \begin{subfigure}[b]{0.85\linewidth}
        \includegraphics[width=\linewidth]{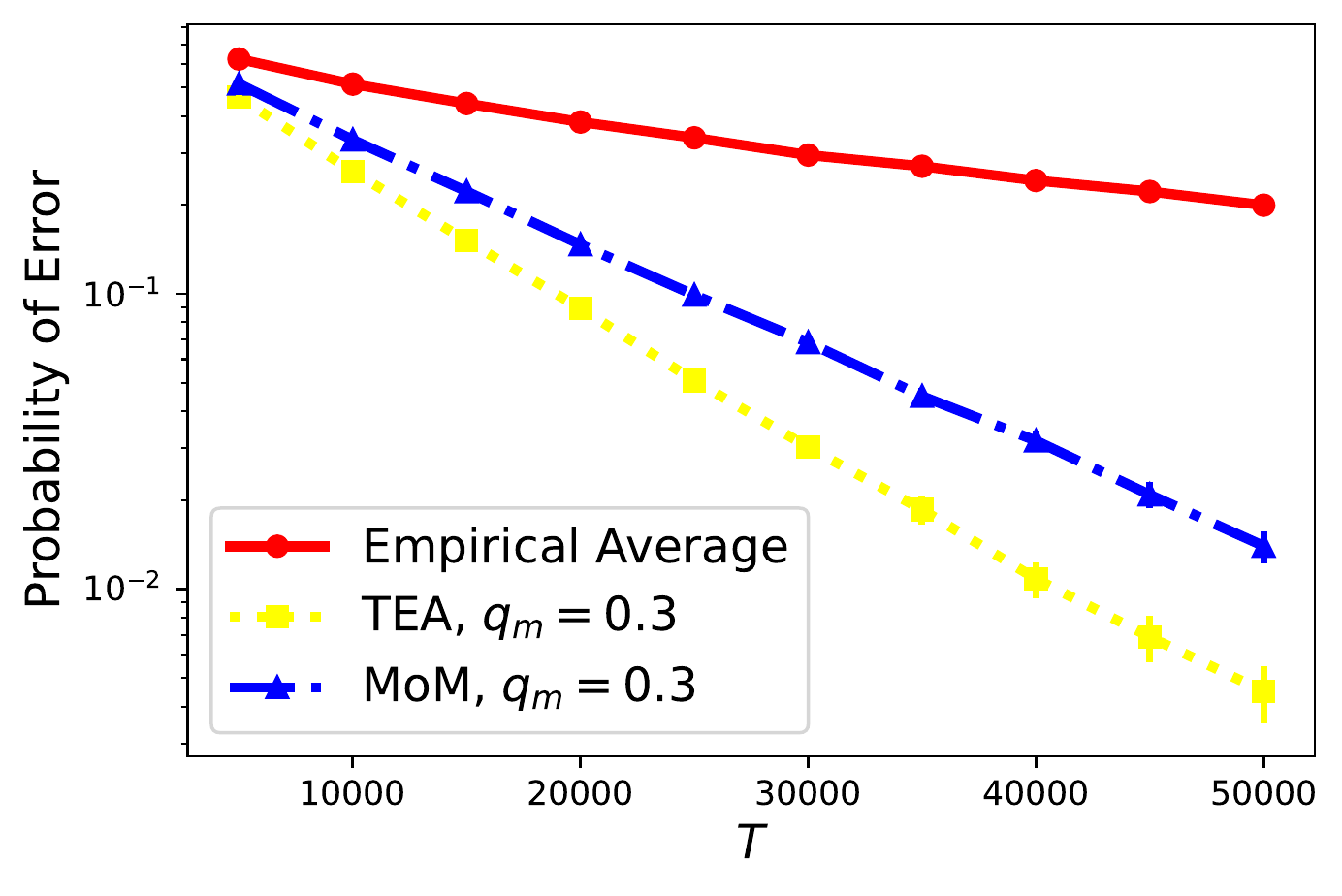}
        \caption{Mean Minimization}
        \label{fig:lomax_mean}
    \end{subfigure} \\
    \begin{subfigure}[b]{0.85\linewidth}
        \includegraphics[width=\linewidth]{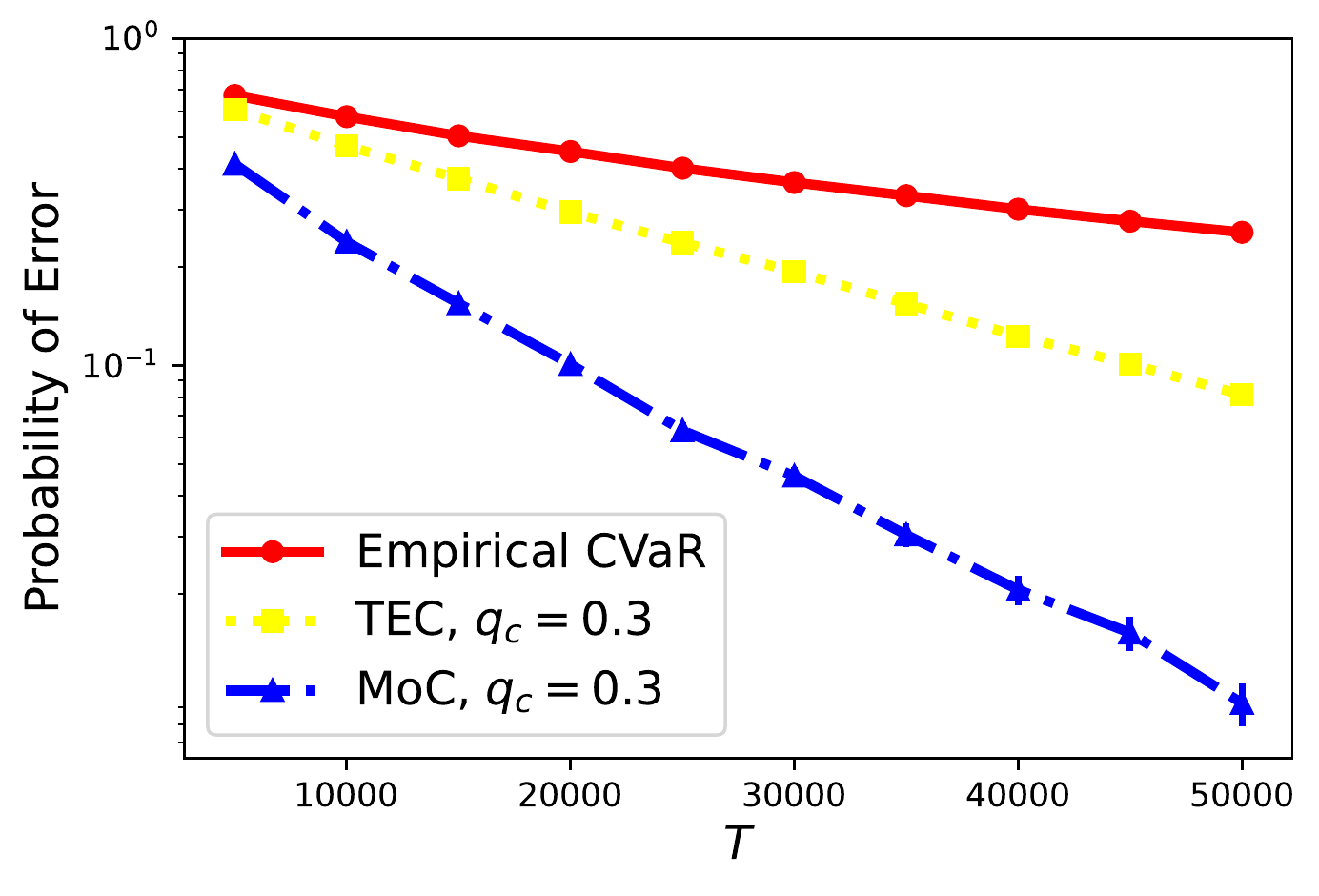}
        \caption{CVaR Minimization}
        \label{fig:lomax_cvar}
    \end{subfigure}
    \label{fig:lomax_case}
    \caption{Lomax Distributed Arms}
\end{figure}

\noindent {\bf Heavy-tailed arms:} Next, consider the case when all
the arms are heavy-tailed. For mean minimization we consider the
following MAB problem instance: there are 10 arms, distributed
according to the lomax distribution
\revisiontwo{
\footnote{Lomax distribution with mean $\mu$ and shape parameter $\gamma>1$ 
is $1- \left(1 + \nicefrac{x}{\left(\mu (\gamma-1)\right)}\right)^{-\gamma}$ for $x>0$ and 0 otherwise}} 
(shape parameter = 1.8), the
optimal arm having mean loss 0.9, and the remaining arms having mean
loss 1.  For CVaR minimization, consider the following MAB problem
instance: there are 10 arms, distributed according to lomax
distribution (shape parameter = 2.0), the optimal having a CVaR 2.55,
and the remaining having a CVaR 3.00.  Note that like the previous
case, parameters $q_m$ and $q_c$ for the truncation estimators are set
to 0.3 and parameters $q_m$ and $q_c$ for the median-of-bins
estimators are also set to 0.3 (see
Sections~\ref{sec:truncated},~\ref{sec:median}).

As can be seen in Figure~\ref{fig:lomax_mean} and
Figure~\ref{fig:lomax_cvar}, the algorithms using empirical estimators
have a markedly inferior performance compared to the truncation-based
and algorithms based on median-of-bins. This is to be expected, since the
latter approaches are more robust to the outliers inherent in
heavy-tailed data.



\noindent {\bf Fragility of specialized algorithms:}
Finally, we present an example where specialized algorithms using
noisy information perform poorly, but our robust algorithms perform
very well. We consider the problem of CVaR minimization on the
following instance involving both heavy-tailed as well as light-tailed
arms: there are 10 arms, five distributed according to a lomax
distribution (shape parameter=2.0; and the CVaR=3.00), and five
distributed exponentially, the optimal arm having a CVaR 2.55, and the
remaining four arms having a CVaR 3.00. As before, we set the
confidence value to be 0.95. What makes this instance (with a
light-tailed optimal arm) challenging is that the bias of
truncation-based estimators can result in a significant
under-estimation of the CVaR for heavy-tailed arms, causing our
algorithms to erroneously declare one of the heavy-tailed arms as optimal.

\begin{figure}[t]
\centering
\includegraphics[width=0.45\textwidth]{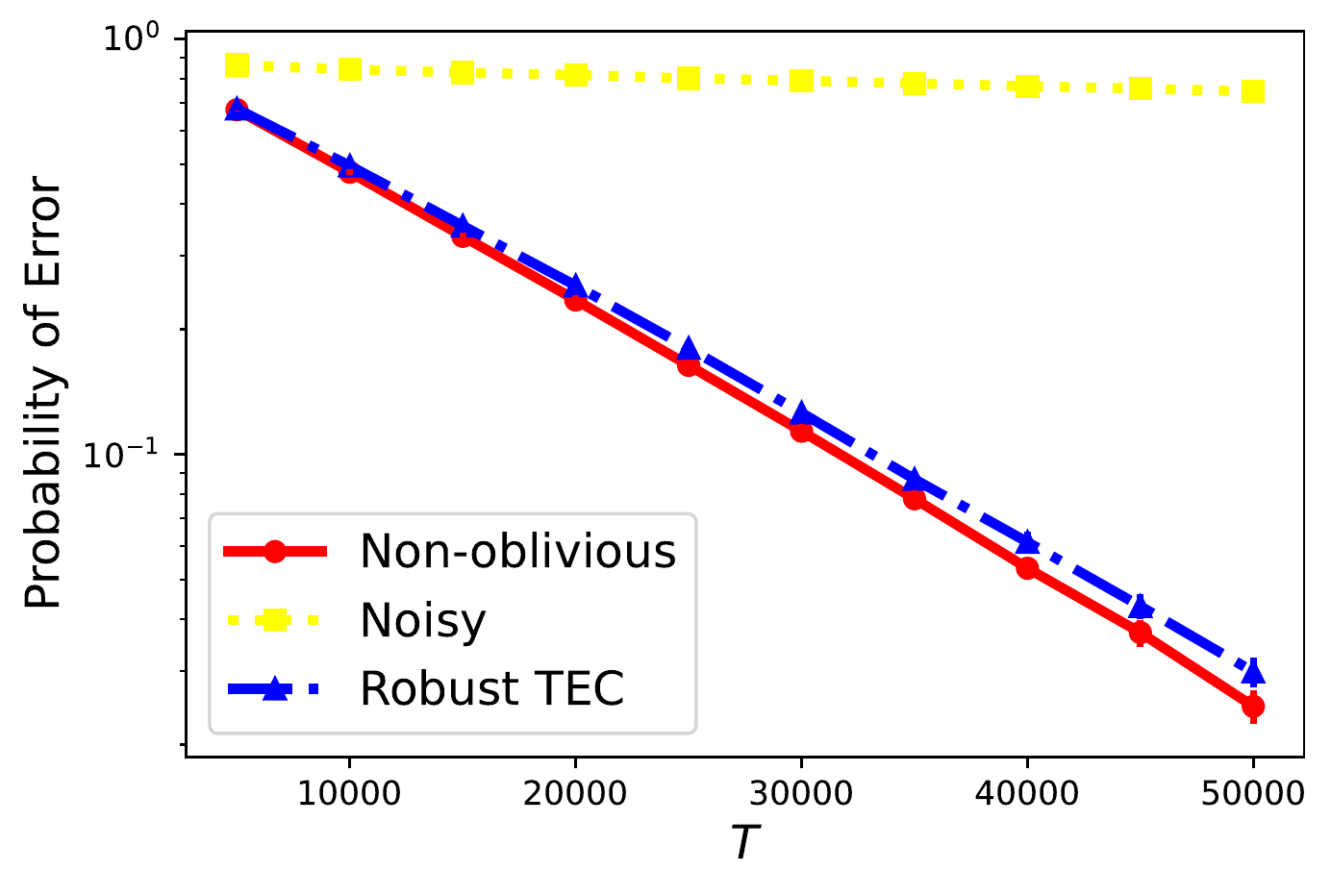}
\caption{Robustness against noisy parameters}
\label{fig:robustness}
\end{figure}

On this instance, we compare the performance of the following
algorithms. Our first candidate is a specialized algorithm that has
access to valid bounds pertaining to the instance. In particular, it
knows $p=1.9,$ $B=0.057,$ and $\Delta=0.45.$ It uses truncated
empirical CVaR as an estimator with truncation parameter being equal
to $(\nicefrac{4B}{\beta \Delta})^{\nicefrac{1}{p-1}}.$ Our second
candidate is another specialized algorithm, but having the following
noisy information, $\hat{p}=2,$ $\hat{B}=0.05,$ and
$\hat{\Delta}=0.6.$ Note that the parameters have been very slightly
perturbed. This algorithm also uses the truncated empirical CVaR as an
estimator with truncation parameter being equal to
$(\nicefrac{4\hat{B}}{\beta \hat{\Delta}})^{\nicefrac{1}{\hat{p}-1}}.$
Our final candidate is a robust algorithm which also has access to the
noisy parameters as stated above but it sets the truncation parameter
as
$(\nicefrac{4\hat{B}}{\beta \hat{\Delta}})^{\nicefrac{1}{\hat{p}-1}} +
T^{0.3}.$  As can be seen in Figure~\ref{fig:robustness}, algorithm 
with noisy estimates performs very poorly but our robust algorithm 
performs nearly as good as the non-oblivious algorithm.

\revisiontwo{
\noindent \textbf{Optimizing a non-trivial combination of mean and CVaR:}
We now present two instances where the objectives are non-trivial convex 
combinations of mean and CVaR. We focus on comparing
the algorithm based on empirical estimators with algoritms based on the 
truncation-based estimators. In the first instance, we compare the 
performance in a reward-seeking setting, i.e., having a lower mean 
(loss) is preferred over having a lower CVaR. 
In the second instance, we compare the algorithms in a risk-averse setting,
i.e., having a lower CVaR is preferred to having a lower mean. 

For the first instance, we set the weight for the mean of an arm, $\xi_1$ to
be equal to 0.9, and we set the weight for the CVaR of an arm, $\xi_2$ to be
equal to 0.1. The optimal arm is Lomax distributed with mean equal to 0.85, 
the shape parameter equal to 2, and the CVaR is approximately 
6.75. The non-optimal arms are also Lomax distributed with their means equal to 1, 
the shape parameters equal to 2.75, and the CVaR is approximately 6.42. 
Notice that the optimal arm has a smaller mean (loss) but a larger 
CVaR when compared to the non-optimal arms. Also, notice that the optimal arm
is more heavy-tailed than the non-optimal arms, owing to the smaller shape 
parameter.  

For the second instance, we set the weight for the mean of an arm, $\xi_1$ to
be equal to 0.1, and we set the weight for the CVaR of an arm, $\xi_2$ to be 
equal to 0.9. The optimal arm is again Lomax distributed with CVaR equal to 
2.55, the shape parameter equal to 2.75, and the mean is approximately 0.40.
The non-optimal arms are also Lomax distributed with CVaR equal to 3, the 
shape parameter equal to 2, and the mean is approximately 0.38. Notice
that the optimal arm has a smaller CVaR but a larger mean compared to the
non-optimal arms. Also, notice that the optimal arm is less heavy-tailed than 
the non-optimal arms, owing to the larger shape parameter. 

The peformance is plotted in Figures~\ref{fig:mean_cvar_rs} and 
\ref{fig:mean_cvar_ra}. We can observe that slower truncation growth
leads to a good performance in the first instance but leads to a 
worse performance in the second instance. The bias of the truncation 
based estimators could lead to significant underestimation of the objective
for more heavy-tailed arms. This is helpful in the first instance
where the optimal arm is more heavy-tailed than the non-optimal arms,
but it is problematic in the second instance where the optimal arm 
is less heavy-tailed than the non-optimal arms.  
The behaviour of algorithms based on median-of-bins 
is qualitatively similar to the truncation-based algorithms but the 
above effect is much more pronounced. In particular, the performance
in the first instance is exceptionally good but the performance in 
the second instance is quite poor. Overall, we observe that 
truncation-based algorithms are more stable than median-of-bins-based
algorithms, although the growth of truncation parameters requires careful
tuning. 

\begin{figure}[t]
    \centering
    \begin{subfigure}[b]{0.9\linewidth}
        \includegraphics[width=\linewidth]{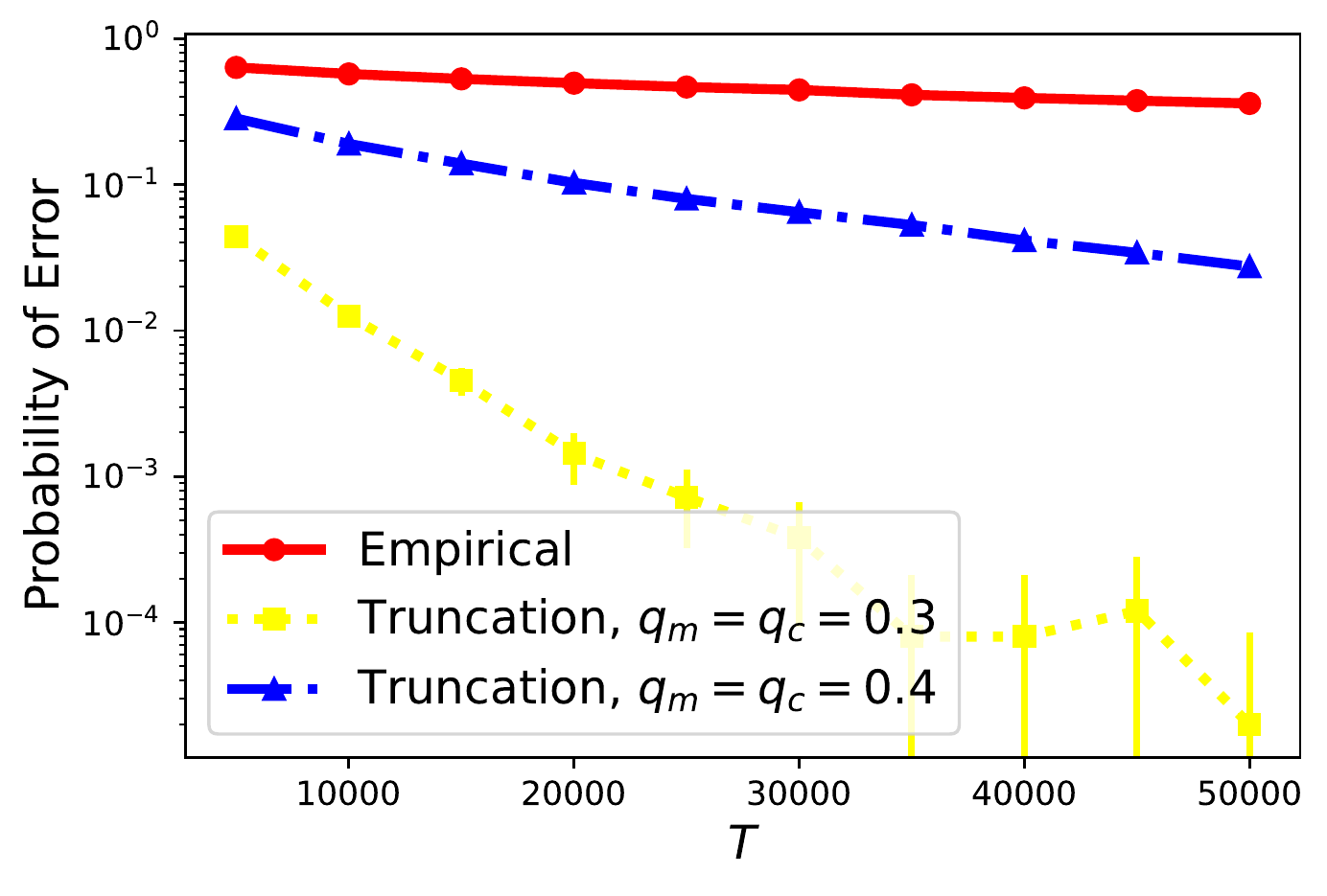}
        \caption{$\xi_1 = 0.9,~\xi_2=0.1$}
        \label{fig:mean_cvar_rs}
    \end{subfigure} \\
    \begin{subfigure}[b]{0.9\linewidth}
        \includegraphics[width=\linewidth]{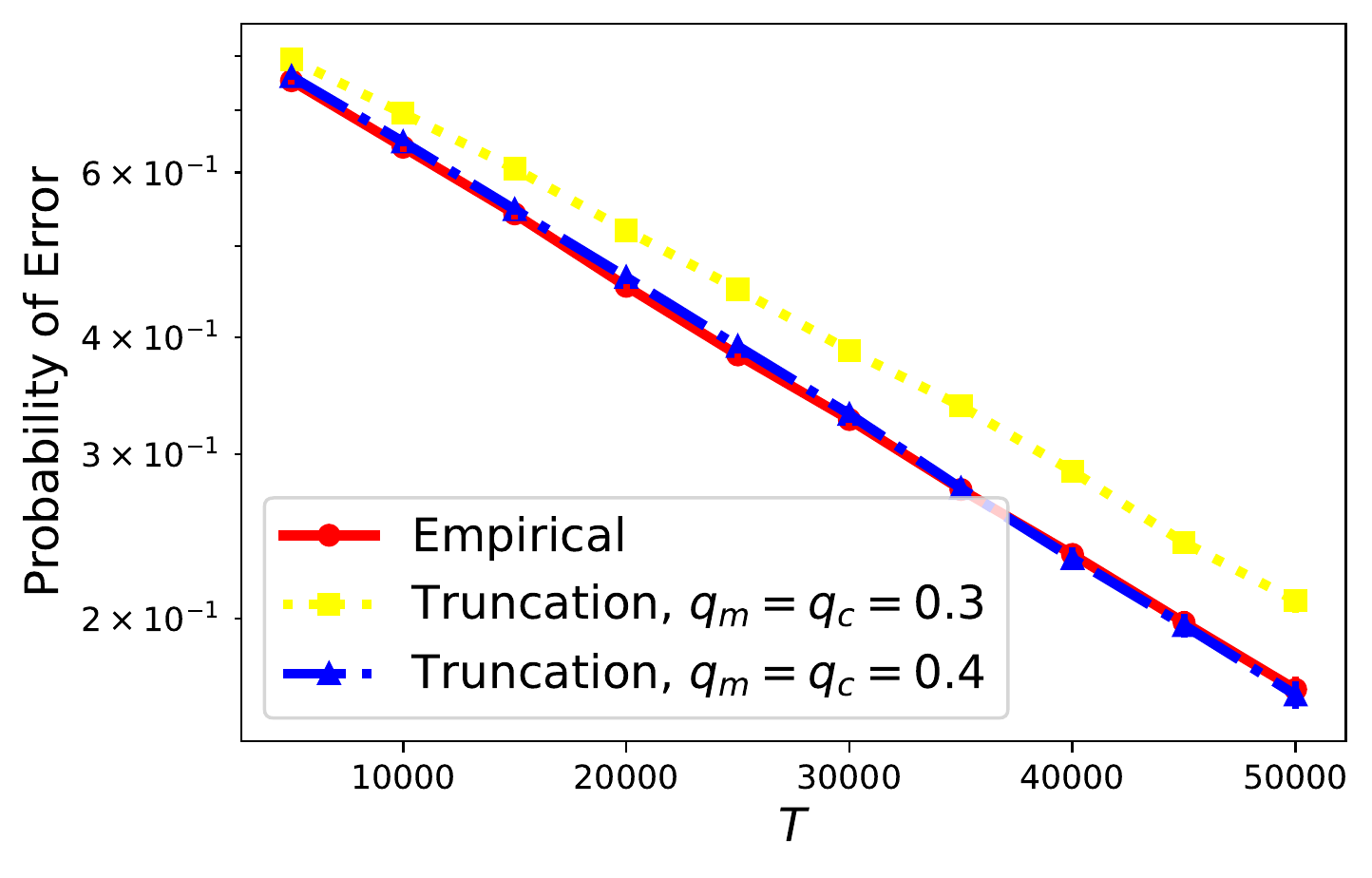}
        \caption{$\xi_1 = 0.1,~\xi_2=0.9$}
        \label{fig:mean_cvar_ra}
    \end{subfigure}
    \label{fig:combination}
    \caption{Non-trivial combination of mean \& CVaR}
\end{figure}

}

\section{Concluding Remarks}
\label{sec:discussion}
\ignore{In this paper, we consider the problem of risk-aware best arm selection in a pure exploration MAB
framework. We study the setting where the algorithms have to
be \textit{distribution oblivious}, i.e., having no knowledge about
the underlying arm distributions. This is in contrast with most
settings in the literature for MAB problems, which assume prior
knowledge of the support, moment bounds, or bounds on the
sub-optimality gaps. We also argue how distribution oblivious
algorithms provide robustness at some loss of performance
guarantees. In particular, we show how blindly applying non-oblivious
approaches to truly oblivious settings might not only degrade
performance but could also lead to inconsistency.

We also established a lower bound on the probability of error for
consistent algorithms operating under the distribution oblivious
setting. It is further shown that naive oblivious algorithms suffer
from a large gap in the performance guarantees when compared to the
lower bound. Our paper proposed two approaches based on the ideas from
robust estimation that match the lower bound on the probability of
error.

The paper motivates future work along several directions. It is
interesting to explore distribution oblivious algorithms in the regret
minimization framework. Moreover, it is not clear which linear
combination of mean and CVaR should be considered for the MAB
problem. Hence, it will be interesting to study risk-constrained MAB
problems in distribution oblivious settings.} 

In this paper, we considered the problem of risk-aware best arm
selection in a pure exploration MAB framework. Our results highlight
the fragility of existing MAB algorithms that require reliable
moment/tail bounds to provide strong performance guarantees. We
established fundamental performance limits of statistically robust MAB
algorithms under fixed budget. We then design algorithms that are
statistically robust to parameter misspecification. Specifically, we
propose distribution oblivious algorithms, i.e., those that do not
need any information on the underlying arms' distributions. The
proposed algorithms leverage ideas from robust statistics and enjoy
near-optimal performance guarantees.

The paper motivates future work along several
directions.
\revision{
First, it would be interesting to design statistically robust variants
of the (stronger) concentration inequalities developed recently for
CVaR and mean in \cite{agrawal2021} and \cite{bourel2020} and apply
those to the MAB problem considered here. Second, it is not always
clear \emph{which} linear combination of mean and CVaR should be
defined as the arm objective in our MAB formulation, given that this
involves expressing the mean cost/reward of an arm and the associated
risk on the same scale. This motivates the analysis of
\emph{risk-constrained} MAB formulations, where the optimal arm is defined as
the one with optimizes the mean cost/reward, subject to a risk
constraint.}

\appendices
\section{Proof of Theorem~\ref{lower-bound-non-obl-theorem}}
\label{app:lower-bound-non-obl}
\revision{
The proof is an easy application of Lemma~\ref{consistency-lemma}
which we will state here again for easy reference.\\ Let $\nu =
(\nu_1, \nu_2)$ be a two-armed bandit model such that $\xi_1 \mu(1)
+ \xi_2 \calp(1) < \xi_1 \mu(2) + \xi_2 \calp(2)$ for given
$\xi_1, \xi_2 \geq 0.$ Any consistent algorithm satisfies
\begin{equation}
\label{eq:lower_bound}
	\limsup_{t \to \infty} -\frac{1}{t} \log p_e(\nu, t) \leq c^*(\nu), 
\end{equation}
where, 
\begin{equation*}
	c^*(\nu) := \inf_{(\nu_1', \nu_2') \in \mathcal{M}: \obj'(1)>\obj'(2)}
	\max(\text{KL}(\nu_1', \nu_1), \text{KL}(\nu_2', \nu_2))
\end{equation*}	  

By taking $(\nu_1', \nu_2') = (\nu_2, \nu_1),$ we get a trivial upper bound
on $c^*(\nu),$ i.e., $c^*(\nu) \leq \max(\text{KL}(\nu_2, \nu_1), \text{KL}(\nu_1, \nu_2)).$
This proves Theorem~\ref{lower-bound-non-obl-theorem}.

}

\section{Proof of Theorem~\ref{conv-cvar-ht-theorem}}
\label{app:conv-cvar-ht-proof}
\begin{theorem}
\label{cvar-conv-ht-true-theorem}
Suppose that $\{X_{i}\}_{i=1}^{n}$ are IID samples distributed as
$X,$ where $X$ satisfies condition {\bf C1}. For $p \in (1,2],$ given $\Delta>0$,  
\begin{subequations}
\begin{align}
  &\begin{aligned}
    &\pbb(\calpn(X) \leq \calp(X) - \Delta) \leq~ \frac{180\revision{V_{\text{emp}}}}{(n\beta)^{p-1}\Delta^p} \\
    &\qquad\qquad\qquad + \text{exp}\Big(-\frac{n\beta}{8}\min\Big(1,\frac{\Delta^{2}\beta^{2/p}}{B^{2/p}}\Big) \Big) 
  \end{aligned} \label{eq:cvar-ht-theorem-a}\\
  &\begin{aligned}
    &\pbb(\calpn(X) \geq \calp(X) + \Delta)\leq~ \frac{360\revision{V_{\text{emp}}}}{(n\beta)^{p-1}\Delta^p} 
    + \frac{72\revision{V_{\text{emp}}}\beta}{(n\beta)^{p-1}B} \\
    &\qquad+\text{exp} \Big(-\frac{n\beta^{1+2/p}\Delta^{2}}{8B^{2/p} + 2\Delta (B\beta)^{1/p}} \Big) 
    + \text{exp} \Big(- \frac{n\beta}{8} \Big)
  \end{aligned}\label{eq:cvar-ht-theorem-b}\\
  &\begin{aligned}
  \text{where } \revision{V_{\text{emp}}} = \frac{2^{p-1}V}{\beta} + 2^p\frac{B}{\beta}.
  \end{aligned}\label{eq:bound_vhat}
\end{align} 
\end{subequations}
\end{theorem}


We will first state three lemmas that will be used repeatedly for proving 
Theorem~\ref{cvar-conv-ht-true-theorem}. We begin by stating a concentration
inequality for empirical average (see Lemma~2, \cite{vakili2013}) . 
\begin{lemma}
\label{ea-conc-lemma}
Let $X$ be a random variable satisfying \textbf{C1}. Let $\hat{\mu}_n$ 
be the empirical mean, then for any $\Delta>0$ we have:
\begin{align*}
	\pbb(|\hat{\mu}_n - \mu| > \Delta) \leq 
  \begin{cases}
    \frac{C_{p} V}{n^{p-1} \Delta^{p}} & \text{for } 1<p\leq 2 \\
    \frac{C_{p} V}{n^{p/2} \Delta^{p}} & \text{for } p > 2
  \end{cases}
\end{align*}	
where $C_{p} = (3\sqrt{2})^{p} p^{p/2}.$
\end{lemma}

Next, consider the inequalities bounding the empirical CVaR estimator 
(see Lemma~3.1, \cite{wang2010}).
\begin{lemma}
\label{wang2010-lemma3}
    Let $X_{[i]}$ be the decreasing order statistics of $X_{i}$; then $f(k) = \frac{1}{k} \sum_{i=1}^{k}X_{[i]}, ~1 \leq k \leq n$, is decreasing and the following two inequalities hold:
    \begin{subequations}
    \begin{align}
        \frac{1}{n\beta} \sum_{i=1}^{\fnb} X_{[i]} &\leq \calpn(X) \leq \frac{1}{n\beta} \sum_{i=1}^{\cnb} X_{[i]} \label{eq:wang-a} \\
        f(\cnb) &\leq \calpn(X) \leq f(\fnb) \label{eq:wang-b}
    \end{align}
    \end{subequations}  
\end{lemma}

We also state the Chernoff Bound for Bernoulli experiments.
\begin{lemma}
\label{lem:chernoff_bound}
Let $Y_{1},...,Y_{n}$ be independent Bernoulli experiments, $\pbb(Y_{i} = 1) = p_{i}$. Set $S = \sum_{i=1}^{n}Y_{i}$, $\mu = \mathbb{E}[Y]$. Then for every $0<\delta<1$, 
\begin{align*}
  P(S \leq (1-\delta)\mu) \leq \text{exp}(-\frac{\mu\delta^{2}}{2}),
\end{align*}
for every $\delta>0$,
\begin{align*}
  P(S \geq (1+\delta)\mu) \leq \text{exp}(-\frac{\mu\delta^{2}}{2+\delta}).
\end{align*}
\end{lemma}

Now, we upper bound the CVaR and mean in terms of parameters $B,$ $p,$ and $\beta.$
\begin{align*}
    \calp(X) =~ &\frac{1}{\beta} \Exp{X\Ind{X\geq\valp(X)}} \\
    =~ &\frac{1}{\beta} \int_{\valp(X)}^{\infty} x dF_{X}(x) \\
    \leq~ & \int_{\valp(X)}^{\infty} |x| \frac{dF_{X}(x)}{\beta} \\
    \leq~ & \Big(\int_{\valp(X)}^{\infty} |x|^{p} \frac{dF_{X}(x)}{\beta} \Big)^{\frac{1}{p}} ~(\text{Using Jensen's Inequality}) \\
    \leq~ & \Big(\int_{-\infty}^{\infty} |x|^{p} \frac{dF_{X}(x)}{\beta} \Big)^{\frac{1}{p}} \\
    \leq~ &\Big(\frac{B}{\beta}\Big)^{\frac{1}{p}} ~(\text{Using bound on } p^{th} \text{ moment}) 
\end{align*}
Similarly, we can show $$\Exp{|X|} \leq \Big(\frac{B}{\beta}\Big)^{\frac{1}{p}}.$$ 
Hence,
\begin{align}
  \calp(X) \leq \Big(\frac{B}{\beta}\Big)^{\frac{1}{p}} \label{eq:cvar-mag-bound} \\
  \Exp{|X|} \leq \Big(\frac{B}{\beta}\Big)^{\frac{1}{p}} \label{eq:mean-mag-bound}
\end{align}

Now, consider the random variable $\Tilde{X}$ which is distributed according to $\pbb(X \in \cdot ~| X \in [\valp(X), \infty))$. Note that $\Exp{\Tilde{X}} = \calp(X)$ and $dF_{\Tilde{X}}(x) = \frac{dF_{X}(x)}{\beta}$. Let us find a bound on $\Exp{|\Tilde{X} - \calp(X)|^{p}}$.
\begin{align*}
  &\Exp{|\Tilde{X} - \calp(X)|^{p}} = \int_{\valp(X)}^{\infty} |x - \calp(X)|^{p} \frac{dF_{X}(x)}{\beta} \\
  \begin{split}
  & \leq \int_{\valp(X)}^{\infty} 2^{p-1} (|x - \mu|^p + |\calp(X) - \mu|^{p}) \frac{dF_{X}(x)}{\beta} \\
  &  \qquad\qquad\qquad\qquad\qquad\qquad (\text{Using Jensen's Inequality})
  \end{split} \\
  & \leq \frac{2^{p-1}V}{\beta} + 2^{p-1}(\calp(X) - \mu)^p \\
  & \leq \frac{2^{p-1}V}{\beta} + 2^{p} \frac{B}{\beta} = \revision{V_{\text{emp}}} 
  \quad (\text{Using }\ref{eq:cvar-mag-bound},~\ref{eq:mean-mag-bound}, \text{ and } \ref{eq:bound_vhat}) \\
\end{align*}
Hence, 
\begin{equation}
  \label{eq:cvar-centred-bound}
  \Exp{|\Tilde{X} - \calp(X)|^{p}} \leq \revision{V_{\text{emp}}}
\end{equation}


\subsection{Proof of \ref{eq:cvar-ht-theorem-a}}
\label{cvar-ht-proof-a}

Let $X_{[i]}$ be the decreasing order statistics of $X_{i}$. We'll condition the probability above on a random variable $\knb$ which is defined as $\knb$ = max$\{i: X_{[i]} \in [\valp(X), \infty)\}$. Note that $\valp(X)$ is a constant such that the probability of a $X$ being greater than $\valp(X)$ is $\beta$. Also observe that $\pbb(\knb = k) = \pbb(k$ from $\{X_{i}\}_{i=1}^{n}$ have values in $[\valp(X), \infty))$. Using the above two statements one can easily see that $\knb$ follows a binomial distribution with parameters $n$ and $\beta$. For ease of notation, we let $p' := \min(p/2, p-1)$.

Consider $k$ IID random variables $\{\Tilde{X}_{i}\}_{i=1}^{k}$ which are distributed according to $\pbb(X \in \cdot ~| X \in [\valp(X), \infty))$.  By conditioning on $\knb = k$, one can observe using symmetry that $\frac{1}{k}\sum_{i=1}^{k} X_{[i]}$ and $\frac{1}{k}\sum_{i=1}^{k}\Tilde{X}_{i}$ have the same distribution. We'll next bound the probability  $\pbb(\calpn(X) \leq \calp(X) - \Delta | \knb = k)$ for different values of $k$. Now,
\begin{align*}
    &\pbb(\calpn(X) \leq \calp(X) - \Delta) \\
    = &\sum_{k=0}^{n}\pbb(\knb = k)\pbb(A) \\
    \leq &\underbrace{\sum_{k=0}^{\fnb}\pbb(\knb = k)\pbb(A)}_{I_{2}}  + \underbrace{\sum_{k=\cnb}^{n}\pbb(\knb = k)\pbb(A)}_{I_{1}}
\end{align*}
where $\pbb(A) = \pbb(\calpn(X) \leq \calp(X) - \Delta | \knb = k)$.

\vspace{3mm}
\textbf{Bounding $I_{1}$}

Note that $k \geq \cnb$. We'll begin by bounding $P(A)$.
\begin{align*}
    &\pbb(\calpn(X) \leq \calp(X) - \Delta | \knb = k) \\
    \leq~ &\pbb \bigg( \frac{1}{\cnb} \sum_{i=1}^{\cnb} X_{[i]}\leq \calp(X) - \Delta | \knb = k \bigg) ~(\text{using \ref{eq:wang-b}}) \\
    \begin{split}
    \leq~ &\pbb \bigg(\frac{1}{k} \sum_{i=1}^{k} X_{[i]} \leq \calp(X) - \Delta | \knb = k \bigg) \\
    & \qquad\qquad\qquad\qquad\qquad\qquad\qquad (\because f(\cdot) \text{ is decreasing})
    \end{split} \\
     =~ &\pbb \bigg(\frac{1}{k} \sum_{i=1}^{k} \Tilde{X}_{i} \leq \calp(X) - \Delta \bigg) \\
    \begin{split}
    \leq~ &\frac{C_{p}\revision{V_{\text{emp}}}}{k^{p'}\Delta^{p}} \quad (\text{Using Lemma~\ref{ea-conc-lemma} \& \eqref{eq:cvar-centred-bound} and } p'=\min(p-1,p/2)) \\
    \end{split}
\end{align*}
Hence, we have the following:
\begin{align*}
	I_{1} =~ &\sum_{k=\cnb}^{n} \binom{n}{k} \beta^k (1-\beta)^{n-k} \pbb(A) \\
	      \leq~ &\sum_{k=\cnb}^{n} \binom{n}{k} \beta^k (1-\beta)^{n-k} \frac{C_{p}\revision{V_{\text{emp}}}}{k^{p'}\Delta^{p}} \\
	      \leq~ &\frac{C_{p}\revision{V_{\text{emp}}}}{(n\beta)^{p'}\Delta^{p}}
\end{align*}
\vspace{3mm}

\textbf{Bounding $I_{2}$} 

Note that $k \leq \fnb$. We'll again start by bounding $\pbb(A)$. For simplicity of notation, we'll denote the $\Big(\frac{B}{\beta}\Big)^{\frac{1}{p}}$ as $b$. Hence, we have $\calp(X) \leq b$ as shown in \ref{eq:cvar-mag-bound}.

\begin{align*}
  &\pbb(\calpn(X) \leq \calp(X) - \Delta | \knb = k) \\
  \leq~ &\pbb \Big( \frac{1}{n\beta} \sum_{i=1}^{\fnb} X_{[i]} \leq \calp(X) - \Delta \Big| \knb = k \Big) ~(\text{Using \ref{eq:wang-a}}) \\
  \begin{split}
  \leq~ &\pbb \Big( \frac{1}{k} \sum_{i=1}^{k} X_{[i]} \leq \frac{n\beta}{k}(\calp(X) - \Delta) \Big| \knb = k \Big) 
  \\ &\qquad\qquad\qquad\qquad\qquad\qquad\qquad\qquad (\because~k\leq \fnb)
  \end{split} \\
  \begin{split}
  \leq~ &\pbb \bigg( \frac{1}{k} \sum_{i=1}^{k} X_{[i]} \leq \calp(X) + \Big(\frac{n\beta}{k} - 1\Big)b  - \frac{n\beta\Delta}{k} \Big| \knb = k \bigg) \\
  &\qquad\qquad\qquad\qquad\qquad\qquad\qquad\qquad (\because \calp(X) \leq b)
  \end{split}
\end{align*}

\textbf{Case 1} $\Delta \in [b,\infty)$

Let $\Delta_{1}(k) = \frac{n\beta\Delta}{k} + \Big(1 - \frac{n\beta}{k}\Big)b = b \bigg(1 + \Big(\frac{\Delta}{b} - 1 \Big)\frac{n\beta}{k} \bigg)$. Note that $\Delta_{1}(k) > 0$ for all $k$ as $\Delta \geq b$. Also note that $\Delta_{1}(k)$ decreases as $k$ increases. As $k \leq n\beta$, $\Delta_{1}(k) \geq \Delta$.

\begin{align*}
	&\pbb \Big(\frac{1}{k} \sum_{i=1}^{k} X_{[i]} \leq \calp(X) - \Delta_{1}(k) | \knb = k \Big)\\
  =~ &\pbb \Big(\frac{1}{k} \sum_{i=1}^{k} \Tilde{X}_{i} \leq \calp(X) - \Delta_{1}(k) \Big) \\
	\leq~ & \frac{C_{p}\revision{V_{\text{emp}}}}{k^{p'}\Delta_{1}^{p}(k)} \\
	\leq~ & \frac{C_{p}\revision{V_{\text{emp}}}}{k^{p'}\Delta^{p}} 
\end{align*}

Now, let us bound $I_{2}$.
\begin{align*}
	I_{2} =~ &\sum_{k=0}^{\fnb} \binom{n}{k}\beta^{k}(1-\beta)^{n-k} \pbb(A) \\
  \begin{split}
  \leq~ &\sum_{k=0}^{\floor{n\beta/2}} \binom{n}{k}\beta^{k}(1-\beta)^{n-k} \\
  &+ \sum_{k=\ceil{n\beta/2}}^{\fnb} \binom{n}{k}\beta^{k}(1-\beta)^{n-k} \frac{C_{p}\revision{V_{\text{emp}}}}{k^{p'}\Delta^{p}}
  \end{split} \\
	\leq~ &\pbb(\knb \leq \floor{n\beta/2}) + \frac{2^{p'}C_{p}\revision{V_{\text{emp}}}}{(n\beta)^{p'}\Delta^{p}} \\
	\leq~ & \frac{2^{p'}C_{p}\revision{V_{\text{emp}}}}{(n\beta)^{p'}\Delta^{p}} + e^{-n\beta/8} \quad (\text{Using Chernoff on }\knb)
\end{align*}

\textbf{Case 2} $\Delta \in (0,b)$

Here, $\Delta_{1}(k) = \frac{n\beta\Delta}{k} - \Big(\frac{n\beta}{k}-1 \Big)b = b \bigg(1 - \Big(1- \frac{\Delta}{b}  \Big)\frac{n\beta}{k} \bigg)$. Note that $\Delta_{1}(k)>0$ iff $k>n\beta(1-\frac{\Delta}{b})$.

\textbf{Case 2.1} If $\Delta$ is very small such that $\fnb \leq n\beta\Big(1-\frac{\Delta}{b}\Big)$, then $\Delta_{1}(k)\leq 0$. Let's bound $I_{2}$ for this case:
\begin{align*}
    I_{2} &\leq \sum_{k=0}^{\fnb} \pbb(\knb = k) \\
    &= \pbb(\knb \leq \fnb) \\
    &\leq \pbb(\knb \leq n\beta(1-\Delta/b)) \\
    & \leq \text{exp} \Big(-n\beta \frac{\Delta^{2}}{2b^{2}} \Big) \qquad (\text{Chernoff on }\knb)
\end{align*}

\textbf{Case 2.2} $n\beta(1-\Delta/b) < \fnb$

Choose $k_{\gamma}^{*} = n\beta(1-\gamma \Delta/b)$ for some $\gamma \in [0,1]$. Then, $n\beta(1-\Delta/b) \leq k_{\gamma}^{*} \leq n\beta$. 

Assume $k^{*}_{\gamma} <\fnb$. The proof can can be easily adapted when $k^{*}_{\gamma} \geq \fnb$. As we will see, the bound on $I_{2}$ is looser when $k^{*}_{\gamma} < \fnb$.

For $k>k_{\gamma}^{*}$, $\Delta_{1}(k)>0$. As $k$ increases, $\Delta_{1}(k)$ also increases. 

Now, we'll bound $\pbb \Big(\frac{1}{k} \sum_{i=1}^{k} X_{[i]} \leq \calp(X) - \Delta_{1}(k) | \knb = k \Big)$:
\begin{align*}
   &\pbb \Big(\frac{1}{k} \sum_{i=1}^{k} X_{[i]} \leq \calp(X) - \Delta_{1}(k) | \knb = k \Big) \\
   = &\pbb \Big(\frac{1}{k} \sum_{i=1}^{k} \Tilde{X}_{i} \leq \calp(X) - \Delta_{1}(k) \Big) \\ 
   \leq & 
   \begin{cases}
	   \frac{C_{p}\revision{V_{\text{emp}}}}{k^{p'}\Delta_1(k)^p}; \ksg < k \leq \fnb \\
	   1 ; \qquad \qquad\qquad\qquad k \leq \ksg
	\end{cases}	\\
   \leq & 
   \begin{cases}
   		\frac{C_{p}\revision{V_{\text{emp}}}(1-\gamma\Delta/b)^{p}}{k^{p'}\Delta^p(1-\gamma)^p}; \ksg < k \leq \fnb \\
	   	1 ; \qquad \qquad\qquad\qquad k \leq \ksg
   \end{cases}
\end{align*}

We will bound $I_2$ as follows:
\begin{align*}
	I_2 =~ &\sum_{k=0}^{\fnb} \binom{n}{k} \beta^k (1-\beta)^{n-k} \pbb(A) \\
	\leq~ &\underbrace{\sum_{k=0}^{\floor{\ksg}} \binom{n}{k}\beta^{k}(1-\beta)^{n-k}}_{I_{2,a}} \\
  + &\underbrace{\sum_{k=\ceil{\ksg}}^{\fnb}\binom{n}{k}\beta^{k}(1-\beta)^{n-k} \frac{C_{p}\revision{V_{\text{emp}}}(1-\gamma\Delta/b)^{p}}{k^{p'}\Delta^p(1-\gamma)^p}}_{I_{2,b}} \\
\end{align*}

Let's bound $I_{2,a}$. This is very similar to \textit{Case 2.1}.
\begin{align*}
    I_{2,a} & = \sum_{k=0}^{\floor{\ksg}} \binom{n}{k}\beta^{k}(1-\beta)^{n-k} \\
    & \leq \pbb\big(\knb \leq (1 - \gamma\Delta/b)n\beta \big) \\
    & \leq \text{exp} \Big(-n\beta \frac{(\gamma\Delta)^{2}}{2b^{2}} \Big)
\end{align*}

If $\ceil{k^{*}_{\gamma}} > \fnb$, $I_{2,b}=0$. 

When $\ceil{k^{*}_{\gamma}} \leq \fnb$, let's bound $I_{2,b}$. 
\begin{align*}
	I_{2,b} \leq~ &\frac{C_{p}\revision{V_{\text{emp}}}(1-\gamma\Delta/b)^{p}}{(n\beta(1-\gamma\Delta/b))^{p'}\Delta^p(1-\gamma)^p} \\
	\leq~ &\frac{C_{p}\revision{V_{\text{emp}}}}{(n\beta)^{p'}\Delta^p(1-\gamma)^p} \quad (\because \Delta \leq b ~\& ~p>p')
\end{align*}

Taking $\gamma=0.5$, we have:
\begin{align*}
	I_2 \leq \frac{2^{p}C_{p}\revision{V_{\text{emp}}}}{(n\beta)^{p'}\Delta^p} + e^{-n\beta\Delta^{2}/8b^2}
\end{align*}
Clearly, the bound above on $I_2$ is looser than that in \textit{Case 2.1}. Comparing the bound above with that in \textit{Case 1}, we have:
\begin{align*}
  I_2 \leq \frac{2^{p}C_{p}\revision{V_{\text{emp}}}}{(n\beta)^{p'}\Delta^p} + \text{exp}\Big(-\frac{n\beta}{8}\min\Big(1,\frac{\Delta^{2}}{b^2}\Big) \Big) \quad (\because p>p')
\end{align*}

Combining bounds on $I_1$ and $I_2$, we finally have,
\begin{align*}
	I \leq~ &\frac{(2^{p}+1)C_{p}\revision{V_{\text{emp}}}}{(n\beta)^{p'}\Delta^p} + \text{exp}\Big(-\frac{n\beta}{8}\min\Big(1,\frac{\Delta^{2}}{b^2}\Big) \Big)
\end{align*}  
When $p \in (1,2],$ the expression above can be simplified to get 
Equation~\eqref{eq:cvar-ht-theorem-a}.

\subsection{Proof of \ref{eq:cvar-ht-theorem-b}}
\label{cvar-ht-proof-b}
Let's prove the second part of this theorem now which is the inequality \ref{eq:cvar-ht-theorem-b}. We'll again condition on random variable $\knb$. 
Remember that $\knb$ follows a binomial distribution with parameters $n$ and $\beta$.

The random variables $\{\Tilde{X}_{i}\}_{i=1}^{k}$ are distributed according to $\pbb(X \in \cdot ~ | X \in [\valp(X), \infty))$. By conditioning of $\knb = k$ distributions of $\frac{1}{k}\sum_{i=1}^{k} X_{[i]}$ and $\frac{1}{k}\sum_{i=1}^{k} \Tilde{X}_{i}$ are same by symmetry. 

\begin{align*}
    &\pbb(\calpn(X) \geq \calp(X) + \Delta)  \\
    = &\sum_{k=0}^{n} \pbb(\knb = k) \pbb(A) \\
    \leq &\underbrace{\sum_{k=0}^{\fnb}\pbb(\knb = k)\pbb(A)}_{I_{1}}  + \underbrace{\sum_{k=\cnb}^{n}\pbb(\knb = k)\pbb(A)}_{I_{2}}
\end{align*}
where $\pbb(A) = \pbb(\calpn(X) \geq \calp(X) + \Delta | \knb = k)$. 
Notice that $I_{1}$ and $I_{2}$ got interchanged from \ref{cvar-ht-proof-a}

\vspace{3mm}
\textbf{Bounding $I_{1}$}

Note that $k \leq \fnb$. Let's bound $\pbb(A)$ for this case:
\begin{align*}
	&\pbb(\calpn(X) \geq \calp(X) + \Delta | \knb = k) \\
	\leq &\pbb \Big( \frac{1}{\fnb} \sum_{i=0}^{\fnb} X_{[i]} \geq \calp(X) 
	+ \Delta | \knb = k\Big) ~(\text{using \ref{eq:wang-b}} ) \\
  \leq &\pbb \Big(\frac{1}{k} \sum_{i=1}^{k} X_{[i]} 
  \geq \calp(X) + \Delta | \knb  = k \Big) \\
  &\qquad\qquad\qquad\qquad\qquad\qquad\qquad(\because f(\cdot) \text{ is decreasing}) \\
  = &\pbb \Big(\frac{1}{k} \sum_{i=1}^{k} \Tilde{X_{i}} 
  \geq \calp(X) + \Delta \Big) \\
  \leq &\frac{C_{p}\revision{V_{\text{emp}}} }{k^{p'}\Delta^{p}}
\end{align*}
Let's bound $I_{1}$ now:
\begin{align*}
	I_{1} =~ &\sum_{k=0}^{\fnb} \binom{n}{k} \beta^{k} (1-\beta)^{n-k} \pbb(A) \\
	\leq~ &\sum_{k=0}^{\floor{n\beta/2}} \binom{n}{k} \beta^{k} (1-\beta)^{n-k} \\
	+ &\sum_{k=\ceil{n\beta/2}}^{\fnb} \binom{n}{k} \beta^{k} (1-\beta)^{n-k} 
	\frac{C_{p}\revision{V_{\text{emp}}}}{k^{p'}\Delta^{p}} \\
	\leq~ &\frac{2^{p'}C_{p}\revision{V_{\text{emp}}}}{(n\beta)^{p'}\Delta^{p}} + e^{-n\beta/8}
	\quad (\text{Using Chernoff on }\knb)
\end{align*}

\textbf{Bounding $I_{2}$}: 

Note that $k \geq \cnb$. Let's begin by bounding $\pbb(A)$:
\begin{align*}
    &\pbb(\calpn(X) \geq \calp(X) + \Delta | \knb = k) \\
    \leq &\pbb \Big( \frac{1}{n\beta} \sum_{i=1}^{\cnb} X_{[i]} \geq \calp(X) + \Delta | \knb = k \Big) ~(\text{using \ref{eq:wang-a}}) \\
    \leq &\pbb \Big(\frac{1}{n\beta}\sum_{i=1}^{k} X_{[i]} \geq \calp(X) + \Delta | \knb = k \Big) ~(\because k \geq \cnb) \\
    = &\pbb \Big( \frac{1}{k} \sum_{i=1}^{k} X_{[i]} \geq \frac{n\beta}{k} (\calp(X)+\Delta) | \knb = k \Big) \\
    \leq &\pbb \Big( \frac{1}{k} \sum_{i=1}^{k} X_{[i]} \geq \calp(X) + \frac{n\beta\Delta}{k} \\
     &\qquad\qquad\qquad\qquad\qquad\qquad- \Big(1 - \frac{n\beta}{k} \Big)b \Big| \knb = k \Big)
\end{align*}
Let $\Delta_{1}(k) = \frac{n\beta\Delta}{k} - \Big(1 - \frac{n\beta}{k} \Big)b = b \Big( (1+\frac{\Delta}{b})\frac{n\beta}{k} - 1 \Big)$. Notice that $\Delta_{1}(k) \geq 0$ if $k \leq  (1+\frac{\Delta}{b}) n\beta $.

Unlike \ref{cvar-ht-proof-a}, we can consider the entire range 
$\Delta \in [0,\infty)$.

\textbf{Case 1.1} If $\Delta$ is very small such that $(1+\frac{\Delta}{b}) n\beta  \leq \cnb$, then $\Delta_{1}(k) \leq 0$. $\frac{\Delta}{b}$ could be any non-negative real and Chernoff bound ahead is adapted for this fact. 

Let's bound $I_{2}$ in this case:
\begin{align*}
    I_{2} & \leq \sum_{k=\cnb}^{n} \pbb(\knb = k) \\
    & = \pbb(\knb \geq \cnb) \\
    & \leq \pbb\Big(\knb \geq (1 + \Delta/b)n\beta \Big) \\
    & \leq \text{exp} \Big(-n\beta \frac{(\Delta/b)^{2}}{2+\Delta/b} \Big) ~(\text{Chernoff on }\knb)
\end{align*}

\textbf{Case 1.2} $ (1+\frac{\Delta}{b}) n\beta  > \cnb$

We choose $\ksg = (1+\frac{\gamma\Delta}{b})n\beta$ for some $\gamma \in [0,1]$. Note that $ (1+\frac{\Delta}{b}) n\beta \geq \ksg \geq n\beta$. Assume that $k^{*}_{\gamma} > \cnb$. The proof when $k^{*}_{\gamma} \leq \cnb$ easily follows. We'll also see that the bound on $I_{2}$ is looser when $k^{*}_{\gamma} > \cnb$. 

Note that $\Delta_{1}(k)$ decreases as $k$ increases. Now,
\begin{align*}
	&\pbb\Big(\frac{1}{k} \sum_{i=1}^{k} X_{[i]} \geq \calp(X) + \Delta_{1}(k) \Big | \knb = k\Big) \\
  = &\pbb\Big(\frac{1}{k} \sum_{i=1}^{k} \Tilde{X}_{i} \geq \calp(X) + \Delta_{1}(k) \Big) \\
    \leq& 
    \begin{cases}
	    \frac{C_{p}\revision{V_{\text{emp}}}}{k^{p'}\Delta_{1}(k)^{p}} \quad \cnb \leq k < k^{*}_{\gamma} \\
	    1; \qquad \qquad\qquad\qquad k\geq k^{*}_{\gamma}
    \end{cases} \\
    \leq &
    \begin{cases}
    	\frac{C_{p}\revision{V_{\text{emp}}}(1+\gamma\Delta/b)^p}{k^{p'}\Delta^{p}(1-\gamma)^p} \quad \cnb \leq k < k^{*}_{\gamma} \\
	    1; \qquad \qquad\qquad\qquad\qquad k\geq k^{*}_{\gamma}
    \end{cases}
\end{align*}

Now, we'll bound $I_{2}$:
\begin{align*}
  &\begin{aligned}
   I_{2} \leq &\sum_{k=\cnb}^{n} \pbb(\knb = k) \times \\
   & \pbb\Big(\frac{1}{k} \sum_{i=1}^{k} \Tilde{X}_{i} \geq \calp(X) + \Delta_{1}(k) | \knb = k \Big)
  \end{aligned} \\
  &\begin{aligned}
    \leq &\underbrace{\sum_{k = \ceil{\ksg}}^{n} \binom{n}{k} \beta^{k} (1-\beta)^{n-k}}_{I_{2,a}}  \\
      &+ \underbrace{\sum_{k = \cnb}^{\floor{\ksg}} \binom{n}{k}\beta^{k}(1-\beta)^{n-k} \frac{C_{p}\revision{V_{\text{emp}}}(1+\gamma\Delta/b)^p}{k^{p'}\Delta^{p}(1-\gamma)^p}}_{I_{2,b}}
  \end{aligned}
\end{align*}

Let's bound $I_{2,a}$ first. Here $\frac{\gamma \Delta}{b}$ could be any non-negative real and Chernoff bound ahead is adapted for this fact.
\begin{align*}
    I_{2,a}  &\leq \pbb(\knb \geq \ksg) \\
    &\leq \pbb \big(\knb \geq (1+\gamma\Delta/b)n\beta\big) \\
    &\leq \text{exp} \Big(-n\beta \frac{\gamma^{2}(\Delta/b)^{2}}{2 + \gamma\Delta/b} \Big) \quad (\text{Chernoff on } \knb)
\end{align*}

If $\floor{\ksg} < \cnb$, then $I_{2,b} = 0$. 
When $\floor{\ksg} \geq \cnb$, let's bound $I_{2,b}$:
\begin{align*}
	I_{2,b} \leq~ &\frac{C_{p}\revision{V_{\text{emp}}}}{(n\beta)^{p'}(1-\gamma)^p}\Big(\frac{1}{\Delta} + \frac{\gamma}{b}\Big)^{p} \\
	\leq~ &\frac{2^{p-1}C_{p}\revision{V_{\text{emp}}}}{(n\beta)^{p'}\Delta^p(1-\gamma)^p} + \frac{2^{p-1}C_{p}\revision{V_{\text{emp}}}\gamma^p}{(n\beta)^{p'}(1-\gamma)^p b^p} \\
  &\qquad\qquad\qquad\qquad\qquad (\text{Using Jensen's Inequality})
\end{align*}

Putting $\gamma = 0.5$, we have:
\begin{align*}
	I_{2} \leq \frac{2^{2p-1}C_{p}\revision{V_{\text{emp}}}}{(n\beta)^{p'}\Delta^p} + \frac{2^{p-1}C_{p}\revision{V_{\text{emp}}}}{(n\beta)^{p'}b^p} + \text{exp} \Big(-n\beta \frac{(\Delta/b)^2}{8+2(\Delta/b)} \Big)
\end{align*}
Clearly, the bound above is looser than that for \textit{Case 1.1}.

Finally, combining the bounds on $I_1$ and $I_2$, we get
\begin{align*}
  \begin{split}
  I \leq~ &\frac{(2^p+1)2^{p-1}C_{p}\revision{V_{\text{emp}}}}{(n\beta)^{p'}\Delta^p} + \frac{2^{p-1}C_{p}\revision{V_{\text{emp}}}}{(n\beta)^{p'}b^p} \\
  +~&\text{exp} \Big(-n\beta \frac{(\Delta/b)^2}{8+2(\Delta/b)} \Big) + \text{exp} \Big(- \frac{n\beta}{8} \Big) 
  \end{split}
\end{align*}
When $p \in (1,2],$ the expression above can be simplified to get 
Equation~\eqref{eq:cvar-ht-theorem-b}.

\section{Proof of Theorem~\ref{empirical-lower-bound-theorem}}
\label{app:conv-cvar-lower-bound-proof}
Let $\{X_{i} \}_{i=1}^n$ be IID samples of a random variable $X$. $X$ is distributed
according to a pareto distribution with parameters $x_m$ and $a$, i.e., the CCDF of
$X$ is given by $P(X>x) = \frac{x_m^a}{x^a}$ for $x>x_m$ and 1 otherwise. Let the
scale parameter $a>1$. Note that the moments smaller than the $a^{\text{th}}$ moment exist. 
Let $p = a-\epsilon$ where $\epsilon$ is a number greater than but arbitrarily close
to zero. One can check that $p^{\text{th}}$ moment exists.

We're interested to lower bound $\pbb(|\calpn(X) - \calp(X)| > \varepsilon)$.
Note that $\pbb(|\calpn(X) - \calp(X)| > \varepsilon) \geq 
\pbb(\calpn(X) > \calp(X) + \varepsilon).$ We'll focus on lower bounding the 
second probability.

Let $X_{[i]}$ be the decreasing order statistics of $X_{i}$. We'll condition the probability above on a random variable $\knb$ which is defined as $\knb$ = max$\{i: X_{[i]} \in [\valp(X), \infty)\}$. As argued before, $\knb$ follows a binomial distribution with parameters $n$ and $\beta$. 

Consider $k$ IID random variables $\{\Tilde{X}_{i}\}_{i=1}^{k}$ which are distributed according to $\pbb(X \in \cdot ~| X \in [\valp(X), \infty))$.  By conditioning on $\knb = k$, one can observe using symmetry that $\frac{1}{k}\sum_{i=1}^{k} X_{[i]}$ and $\frac{1}{k}\sum_{i=1}^{k}\Tilde{X}_{i}$ have the same distribution. 

Now, we'll lower bound $\pbb(\calpn(X) > \calp(X) + \varepsilon | \knb = k)$
when $k \geq \cnb$
\begin{align*}
	&\pbb(\calpn(X) > \calp(X) + \varepsilon | \knb = k) \\
	\geq &\pbb\left(\frac{1}{\cnb} \sum_{i=1}^{\cnb} X_{[i]} > \calp(X) + \varepsilon | \knb = k \right)\\
	&\qquad\qquad\qquad\qquad\qquad\qquad (\text{using Equation}~\eqref{eq:wang-b}) \\
	\geq &\pbb\left(\frac{1}{k} \sum_{i=1}^{k}X_{[i]} > \calp(X) + \varepsilon | \knb = k \right) \\
	& \qquad\qquad\qquad\qquad\qquad (\because k \geq \cnb \text{ and using Lemma}~\ref{wang2010-lemma3}) \\
	= &\pbb\left( \frac{1}{k} \sum_{i=1}^{k} \Tilde{X}_{i} > \calp(X) + \varepsilon \right) \\
	\geq &\pbb\left( \exists i \in [k] \text{ such that } \Tilde{X}_{i} > k (\calp(X) + \varepsilon) \right) \\
	= &1 - \left(1 - \frac{x_m^a}{k^a(\calp(X)+\varepsilon)^a} \right)^k \\
	\geq &1 - \exp \left(- \frac{x_m^a }{k^{a-1} (\calp(X)+\varepsilon)^a} \right)  
\end{align*}
Hence, we have
\begin{align*}
	&\pbb(\calpn(X) > \calp(X) + \varepsilon)\\
	 \geq~ &\pbb(\knb \geq \cnb) 
	\left(1 - \exp \left(- \frac{x_m^a}{n^{a-1} (\calp(X)+\varepsilon)^a} \right)\right) \\
	\stackrel{(1)}\geq~ &\beta \left(1 - \exp \left(- \frac{x_m^a}{n^{a-1} (\calp(X)+\varepsilon)^a} \right)\right) \\
	=~ &\frac{\beta x_m^a}{n^{a-1} (\calp(X)+\varepsilon)^a} + o\left(\frac{1}{n^{a-1}}\right)
\end{align*}	
Here, $1$ follows because $\pbb(\knb \geq \cnb) \geq \beta.$ See Equation~3 in \cite{pelekis2016}.

\section{Proof of Theorem~\ref{cvar-mob-theorem}}
\label{app:cvar-mob-proof}
The value of $N^*$ is given by
\begin{multline}
\label{eq:mob-threshold}
N^{*} = \max\bigg( \Big(\frac{4320\revision{V_{\text{emp}}}}{\beta^{p-1}\Delta^p} 
+ \frac{576\revision{V_{\text{emp}}}\beta}{\beta^{p-1}B}\Big)^{\frac{1}{p-1}}, \\
\frac{\log(24)}{\beta}\max\Big(8, \frac{8B^{2/p}}{\Delta^2\beta^{2/p}} + \frac{2B^{1/p}}{\Delta\beta^{1/p}} \Big) \bigg),
\end{multline}
where $\revision{V_{\text{emp}}}$ is a constant that depends of $(p,B,V)$ (see Equation~\ref{eq:bound_vhat}).

For each bin $i$ define a random variable $Y_{i} = \Ind{|\hat{c}_{\alpha, N}(i)-\calp(X)|>\Delta}$. $Y_{i}$ takes the value 1 with probability $\hat{p}$. From equation~\ref{eq:cvar-ht-theorem-a} and equation~\ref{eq:cvar-ht-theorem-b}, we have:
\begin{align*}
 	\hat{p} \leq &\frac{540\revision{V_{\text{emp}}}}{(N\beta)^{p-1}\Delta^p} 
    + \frac{72\revision{V_{\text{emp}}}\beta}{(N\beta)^{p-1}B} + \text{exp} \Big(- \frac{N\beta}{8}\Big) \\
 	&+~ \text{exp}\Big(-\frac{N\beta}{8}\min\Big(1,\frac{\Delta^{2}}{b^2}\Big) \Big)  
 	+ \text{exp} \Big(-\frac{N\beta\Delta^{2}}{8b^2 + 2\Delta b} \Big).
 \end{align*} 
where $b = \Big(\frac{B}{b}\Big)^{\frac{1}{p}}$.\\
A sufficient condition to ensure that $\hat{p}$ is less than 0.25 is the following:
\begin{align}
 	&\frac{540\revision{V_{\text{emp}}}}{(N\beta)^{p-1}\Delta^p} 
    + \frac{72\revision{V_{\text{emp}}}\beta}{(N\beta)^{p-1}B} \leq \frac{1}{8} \nonumber \\
 	\Leftarrow &N \geq 
 	\Big(\frac{4320\revision{V_{\text{emp}}}}{\beta^{p-1}\Delta^p} 
    + \frac{576\revision{V_{\text{emp}}}\beta}{\beta^{p-1}B}\Big)^{\frac{1}{p-1}} \label{eq:mob-threshold-1}
 \end{align} 
and 
\begin{align}
	&\text{exp} \Big(- \frac{N\beta}{8} \Big)   
 	+ \text{exp} \Big(-\frac{N\beta\Delta^{2}}{8b^2 + 2\Delta b} \Big) 
 	\nonumber \\
 	&\qquad\qquad\qquad\qquad
 	+ \text{exp}\Big(-\frac{N\beta}{8}\min\Big(1,\frac{\Delta^{2}}{b^2}\Big) \Big) \leq \frac{1}{8} \nonumber \\
 	\Leftarrow &N \geq \frac{\log(24)}{\beta} \max\Big(8, \frac{8b^2}{\Delta^2} + \frac{2b}{\Delta} \Big). \label{eq:mob-threshold-2}
\end{align}
Using Equations~\ref{eq:mob-threshold-1} and \ref{eq:mob-threshold-2}, we get Equation~\ref{eq:mob-threshold}.
\vspace{2ex}

Now, for $N \geq N^{*}$
\begin{align*}
	&\prob{|\hat{c}_{M} - \calp(X)| > \Delta} \\
	\leq~ &\prob{\sum_{i=1}^{k} Y_{i} \geq k/2} \\
	\leq~ &\text{exp}(-2k(0.5-\hat{p})^2) \quad(\text{Using Hoeffding's Inequality}) \\
	\leq~ &\text{exp}(-k/8) \\
	\leq~ &\text{exp}(-\frac{n}{8N})
\end{align*}

\section{Proof of Theorem~\ref{sr-prob-of-error-tea}}
\label{app:sr-truncation}

In Successive Rejects algorithm, all arms are played at least 
$\frac{T-K}{K \overline{\log(K)}}$ times. Hence, the least value that
the truncation parameter for mean takes is 
$\left(\frac{T-K}{K \overline{\log(K)}}\right)^{q_m}$ and the least value
that truncation parameter for CVaR takes is
$\left(\frac{T-K}{K \overline{\log(K)}}\right)^{q_c}.$
$T$ should take a value such that the truncation parameters are large 
enough to for the guarantees to kick in. Using Theorem~\ref{cvar-ht-theorem} 
and Lemma~\ref{tea-conc-lemma} which we prove next, we can easily get
Theorem~\ref{sr-prob-of-error-tea}. 

\begin{lemma}
\label{tea-conc-lemma}
Suppose that $\{X_{i}\}_{i=1}^{n}$ are IID samples distributed as
$X,$ where $X$ satisfies condition {\bf C1}. Given $p \in (1,2]$ and
$\Delta > 0,$
\begin{align}
\prob{|\mu(X) - \tea(X)| \geq \Delta} \leq 2 
\exp\Big(-n^{1-q} \frac{\Delta}{4}\Big) \\
\text{ for } n > \left(\frac{3B}{\Delta}\right)^{1/q(p-1)}.
\end{align}
\end{lemma}

First, consider the following lemma proved in \cite{yu2018} (see Lemma~1).
\begin{lemma}
	\label{tea-conc-lemma}
	Assume that $\{X_{i}\}_{i=1}^{n}$ be $n$ IID samples drawn from the 
	distribution of $X$ which satisfies condition \textbf{C1}. 
	Let $\{b_{i}\}_{i=1}^{n}$ be the truncation parameters for samples 
	$\{X_{i}\}_{i=1}^{n}$. Then, with probability at least $1-\delta$, 
	\begin{align*}
		|\mu(k) - \tea_{n}(k)| \leq
		\begin{cases}
		\frac{\sum_{i=1}^{n} B/b_{i}^{p-1}}{n} +  \frac{2 b_{n} \text{log}(2/\delta)}{n} 
		& \\
		 \qquad\qquad\qquad +\frac{B}{2b_{n}^{p-1}}, &  p \in (1,2]\\
		\frac{\sum_{i=1}^{n} B/b_{i}^{p-1}}{n} 
		+ \frac{2 b_{n} \text{log}(2/\delta)}{n} 
		& \\
		\qquad\qquad\qquad + \frac{B^{2/p}}{2 b_{n}}, & p \in (2,\infty)
		\end{cases}
	\end{align*}
\end{lemma}

All the truncation parameters $\{b_i\}_{i=1}^{n}$ are set to $n^q$ for our algorithm.
We derive bounds for both the cases when $p \in (1,2]$ and $p \in (2,\infty]$. \\ 
\textbf{Case 1} $p \in (1,2]$ \\
Using Lemma~\ref{tea-conc-lemma}, if $p \in (1,2]$
\begin{align*}
	|\mu(k) - \tea_{n}(k)| \leq &\frac{\sum_{i=1}^{n} B/b_{i}^{p-1}}{n} 
	+  \frac{2 b_{n} \log(2/\delta)}{n} + \frac{B}{2b_{n}^{p-1}} \\
	& \leq \frac{3B}{2n^{q(p-1)}} + \frac{2}{n^{1-q}}\log(2/\delta). 
\end{align*}
We want to find $n^{*}$ such that for all $n > n^{*}$:
\begin{align*}
	\underbrace{\frac{3B}{2n^{q(p-1)}}}_{T_{1}} + \underbrace{\frac{2}{n^{1-q}}\log(2/\delta)}_{T_{2}} < \Delta.
\end{align*}
Sufficient condition to ensure the above inequality is to make the $T_{1} < \Delta/2$ and $T_{2} \leq \Delta/2$. $T_{1} \leq \Delta/2$ if
\begin{equation*}
	n > \Big(\frac{3B}{\Delta}\Big)^{\frac{1}{q(p-1)}}.
\end{equation*}
Equating $T_{2} = \Delta/2$, we get
\begin{equation*}
	\delta = 2 \text{exp}\Big(-n^{1-q} \frac{\Delta}{4}\Big).
\end{equation*}

\textbf{Case 2} $p \in (2,\infty)$

Using Lemma~\ref{tea-conc-lemma}, if $p \in (2,\infty)$:
\begin{align*}
	|\mu(k) - \tea_{n}(k)| \leq &\frac{\sum_{i=1}^{n} B/b_{i}^{p-1}}{n} 
	+ \frac{2 b_{n} \log(2/\delta)}{n} + \frac{B^{2/p}}{2 b_{n}} \\
	\leq & \frac{B}{n^{q(p-1)}} + \frac{B}{2n^{q}} +  \frac{2 \log(2/\delta)}{n^{1-q}} \\
	\leq & \frac{3B}{2n^{q}} + \frac{2 \log(2/\delta)}{n^{1-q}}
\end{align*}
We want to find $n^{*}$ such that for all $n > n^{*}$:
\begin{align*}
	\underbrace{\frac{3B}{2n^{q}}}_{T_{1}} + \underbrace{\frac{2 \log(2/\delta)}{n^{1-q}}}_{T_{2}} < \Delta \\
\end{align*}
Sufficient condition to ensure the above inequality is to make the $T_{1} < \Delta/2$ and $T_{2} \leq \Delta/2$.
$T_{1} < \Delta/2$ if:
\begin{equation*}
	n > \Big(\frac{3B}{\Delta}\Big)^{\frac{1}{q}}
\end{equation*}
Equating $T_{2} = \Delta/2$, we get:
\begin{equation*}
	\delta = 2 \text{exp}\Big(-n^{1-q} \frac{\Delta}{4}\Big)
\end{equation*}

\section{Proof of Theorem~\ref{sr-prob-of-error-mob}}
\label{app:sr-mob}
A sufficient condition for Theorem~\ref{sr-prob-of-error-tea} to hold is the following
\begin{align*}
\frac{T-K}{K \overline{\log}(K)} \geq \max &\bigg( \left(\frac{576 \xi_1 V}{\Delta[2]}\right)^{1/q_m}, \left(\frac{8\log(24)}{\beta}\right)^{1/q_c}, \\
&\Big(\frac{4320 \xi_2^p 4^p \revision{V_{\text{emp}}}}{\beta^{p-1}\Delta[2]^p} 
+ \frac{576\revision{V_{\text{emp}}}\beta}{\beta^{p-1}B}\Big)^{\frac{1}{q_c(p-1)}},  \\
& \left(\frac{8\log(24)}{\beta} 
\left( \frac{128 \xi_2^2 B^{2/p}}{\Delta[2]^2\beta^{2/p}} 
+ \frac{8 \xi_2 B^{1/p}}{\Delta[2]\beta^{1/p}}\right)\right)^{1/q_c}\bigg).
\end{align*}
The proof is based on Equation~\ref{eq:mob-threshold} and Lemma~\ref{mom-conc-lemma}
which we prove next.

\begin{lemma}
\label{mom-conc-lemma}
Suppose that $\{X_{i}\}_{i=1}^{n}$ are IID samples distributed as
$X,$ where $X$ satisfies condition {\bf C1}. Let $N = \floor{n/k}$
and $\{\mu_{N}(l)\}_{l=1}^{k}$ be the empirical CVaR estimators for bins
$\{\{X_{j}\}_{j=(l-1)N+1}^{l N}\}_{l=1}^{k}.$ Let $\hat{\mu}_M$ be the median
of empirical CVaR estimators $\{\mu_{N}(l)\}_{l=1}^{k}$, then for $p \in (1,2],$
given $\Delta > 0,$ 
\begin{equation}
 	\prob{|\hat{\mu}_M - \mu(X)| \geq  \Delta} \leq \text{exp}(-\frac{n}{8N})\\
\end{equation} 
for 
\begin{equation}
	N \geq \left(\frac{144V}{\Delta^p}\right)^{1/(p-1)}.
\end{equation}
\end{lemma}
\begin{IEEEproof}
For each bin $l$ define a random variable 
$Y_{l} = \Ind{|\hat{c}_{\alpha, N}(i)-\calp(X)|>\Delta}$. $Y_{l}$ takes the value 1 with probability $\hat{p}$. Using Using Lemma~\ref{ea-conc-lemma},
we have $\hat{p} \leq \frac{C_{p}V}{N^{\min(p-1, p/2)}\Delta^p}.$ 
If we ensure that $\hat{p} \leq 0.25,$ then
\begin{align*}
	&\pbb(|\hat{\mu}_{M} - \mu(X)| > \Delta) \\
	\leq~ &\pbb(\sum_{i=1}^{k} Y_{i} \geq k/2) \\
	\leq~ &\text{exp}(-2k(0.5-\hat{p})^2) \quad(\text{Using Hoeffding's Inequality}) \\
	\leq~ &\text{exp}(-k/8) \\
	\leq~ &\text{exp}(-\frac{n}{8N})
\end{align*}
If $N \geq \left(\frac{4 C_p V}{\Delta^p}\right)^{1/\min(p-1, p/2)},$ then $\hat{p} \leq 0.25.$
Upper bounding $C_p$ when $p \in (1,2]$ gives the statement of the theorem.
\end{IEEEproof}

\section{Bouding magnitude of VaR}
\label{app:var-mag-bound}
Here is an upper bound on the magnitude of $\valp(X)$ in terms of $(p, B, V).$
\begin{lemma}
	\label{mod-var-bound}
	\begin{equation*}
		|\valp(X)| \leq \Big(\frac{B}{\text{min}(\alpha,\beta) } \Big)^{\frac{1}{p}}
	\end{equation*}
\end{lemma}

\begin{IEEEproof}
If $\valp(X)>0$, by definition:
\begin{align*}
	1 - \alpha = &\int_{\valp(X)}^{\infty} dF_{X}(x)  \\
			   = &\int_{\valp(X)}^{\infty} |x|^{p}/|x|^{p} dF_{X}(x) \\
			   \leq &B/|\valp(X)|^{p}
\end{align*}
Hence, $|\valp(X)| \leq (\frac{B}{\beta})^{\frac{1}{p}}$.

If $\valp(X)<0$, by definition:
\begin{align*}
	\alpha = &\int_{-\infty}^{\valp(X)} dF_{X}(x) \\
		   = &\int_{-\infty}^{\valp(X)} |x|^{p}/|x|^{p} dF_{X}(x) \\
		   \leq &B/|\valp(X)|^{p}
\end{align*}
Hence, $|\valp(X)| \leq (\frac{B}{\alpha})^{\frac{1}{p}}$.	
\end{IEEEproof}

\revision{
\section{Proof of Lemma~\ref{lem:chernoff_bound}}
The most accessible proof for the lemma was found in these lecture notes (see \cite{goemans2013}) 
but we will state the proof here for completeness.    

Using Markov's inequality, we have
\begin{align*}
	P(S \geq a) \leq \frac{E[e^{tS}]}{e^{ta}} \text{ and } 
	P(S \leq a) \leq \frac{E[e^{-tS}]}{e^{-ta}}.
\end{align*}
Let us denote the moment generating function (MGF) of $S,$ $E[e^{tS}]$ by $M_{S}(t)$ and the MGF of $Y_i,$ $E[e^{tY_i}]$ by $M_{Y_i}(t).$ As $Y_i$'s are independent, we have
\begin{align*}
	M_S(t) = \prod_{i=1}^n M_{Y_i}(t).
\end{align*}
Upper bounding the the MGF of bernoulli random variables $Y_i$'s in the following manner will be useful for analysis.
\begin{align*}
M_{Y_i}(t) &= p_i e^t + (1-p_i) \\
		   &= 1 + p_i (e^t - 1) \\
		   &\leq e^{p_i(e^t - 1)} \\
		   & (\text{Using } 1 + r \leq e^r \text{ with } r = p_i(e^t - 1)).
\end{align*}
This gives 
\begin{align*}
	M_S(t) &\leq e^{\sum_{i=1}^n p_i (e^t - 1)} \\
	&\leq e^{\mu (e^t - 1)} \quad \because \mu = \sum_{i=1}^n p_i. 
\end{align*}
For the upper tail, we have,
\begin{align*}
	P(S \geq (1+\delta)\mu) &\leq e^{-(1+\delta)\mu t} e^{\mu (e^t - 1)} \\
	&\leq \left(\frac{e^\delta}{(1+\delta)^{1+\delta}} \right)^\mu \\
	&(\text{Maximum is achieved when } t = \log(1+\delta)). 
\end{align*}
We will next take the logarithm of RHS, which is equal to 
\begin{align*}
	\mu(\delta - (1+\delta) \log(1+\delta)).
\end{align*}
We next use the following inequality to upper bound the term above.
\begin{align*}
	\log(1+r) \geq \frac{r}{1 + \nicefrac{r}{2}}.
\end{align*}
The inequality above is easy to show. This gives us
\begin{align*}
 	\mu(\delta - (1+\delta) \log(1+\delta)) \leq \mu \left(\delta  - \frac{\delta(1+\delta)}{1+\nicefrac{\delta}{2}}\right) \leq - \frac{\mu \delta^2}{\delta + 2}.
 \end{align*} 
Hence, we have the following bound on the upper tail 
\begin{align*}
	P(S \geq (1+\delta)\mu) \leq \exp\left(- \frac{\mu \delta^2}{\delta + 2} \right).
\end{align*}

The proof of the lower tail is similar. We put $t = \log(1-\delta)$ and use the following inequality
which holds when $\delta \in (0,1),$
\begin{align*}
	\log(1-\delta) \geq -\delta + \frac{\delta^2}{2}.
\end{align*}

}

\section*{Acknowledgment}
This research was supported in part by the International Centre for
Theoretical Sciences (ICTS) during a visit for the program - Advances
in Applied Probability (Code: ICTS/paap2019/08). We also thank the
anonymous reviewers for their helpful suggestions. 

\ifCLASSOPTIONcaptionsoff
  \newpage
\fi



%
\bibliographystyle{IEEEtran}
\bibliography{references}

%


\begin{IEEEbiographynophoto}{Anmol Kagrecha} received the B.Tech. 
and M.Tech. degrees in electrical engineering (EE) from IIT Bombay 
in 2020, where he was also a recipient of the Institute Silver Medal.    
He is currently a Ph.D. student at the EE department at Stanford 
University, where he is supported by Robert Bosch Stanford Graduate 
Fellowship. His research interests lie in the areas of
reinforcement learning, and information theory. 
\end{IEEEbiographynophoto}

\begin{IEEEbiographynophoto}{Jayakrishnan Nair} 
received the B.Tech. and
M.Tech. degrees in electrical engineering (EE) from
IIT Bombay in 2007 and the Ph.D. degree in EE
from the California Institute of Technology in 2012.
He held post-doctoral positions at the California
Institute of Technology and Centrum Wiskunde \&
Informatica. He is currently an Associate Professor
of EE at IIT Bombay. His research focuses on
modeling, performance evaluation, and design issues
in queueing systems and communication networks.
\end{IEEEbiographynophoto}

\begin{IEEEbiographynophoto}{Krishna Jagannathan} (Member, IEEE) 
received the B.Tech. degree in electrical engineering from 
IIT Madras in 2004, and the S.M. and Ph.D. degrees in 
electrical engineering and computer science from the 
Massachusetts Institute of Technology in 2006 and 2010, 
respectively. He is an Associate Professor with the 
Department of Electrical Engineering, IIT Madras. \
His research interests lie in the areas of stochastic 
modeling and analysis of communication networks, 
online learning, network control, and queueing theory.
\end{IEEEbiographynophoto}






\end{document}